\documentclass[letterpaper]{article} 
\usepackage{aaai2026}  
\usepackage{times}  
\usepackage{helvet}  
\usepackage{courier}  
\usepackage[hyphens]{url}  
\usepackage{graphicx} 
\usepackage{natbib}  
\usepackage{caption} 
\usepackage{algorithm}
\usepackage{algorithmic}

\usepackage{newfloat}
\usepackage{listings}

\usepackage{graphicx}    
\usepackage{adjustbox}   
\usepackage{booktabs}    
\usepackage{multirow}    
\usepackage{makecell}    
\usepackage{array}       
\usepackage{tabularx}    

\usepackage{latexsym}
\usepackage{amsmath,amssymb}
\usepackage{enumitem}
\usepackage{cuted}
\usepackage{textcomp}
\usepackage[T1]{fontenc}      
\usepackage{xcolor}
\usepackage{bm}
\usepackage{amsmath}
\usepackage{subcaption}  

\DeclareCaptionStyle{ruled}{labelfont=normalfont,labelsep=colon,strut=off} 
\floatstyle{ruled}
\newfloat{listing}{tb}{lst}{}
\floatname{listing}{Listing}
\title{DOS: Directional Object Separation in Text Embeddings \\
for Multi-Object Image Generation}

\author{
    Dongnam Byun\textsuperscript{\rm 1,3},
    Jungwon Park\textsuperscript{\rm 1},
    Jungmin Ko\textsuperscript{\rm 2},
    Changin Choi\textsuperscript{\rm 2,4},
    Wonjong Rhee\textsuperscript{\rm 1,2}
}
\affiliations{
    \textsuperscript{\rm 1}Department of Intelligence and Information, Seoul National University\\
    \textsuperscript{\rm 2}Interdisciplinary Program in Artificial Intelligence, Seoul National University\\
    \textsuperscript{\rm 3}Samsung Research, Samsung Electronics\\
    \textsuperscript{\rm 4}SAIT, Samsung Electronics\\
    \{east928, quoded97, jungminko, ci2015.choi, wrhee\}@snu.ac.kr
}

\makeatletter
\let\saved@maketitle\@maketitle
\makeatother

\begin{document}

\maketitle

\begin{figure*}[!t]
  \centering
  \includegraphics[width=0.9\textwidth]{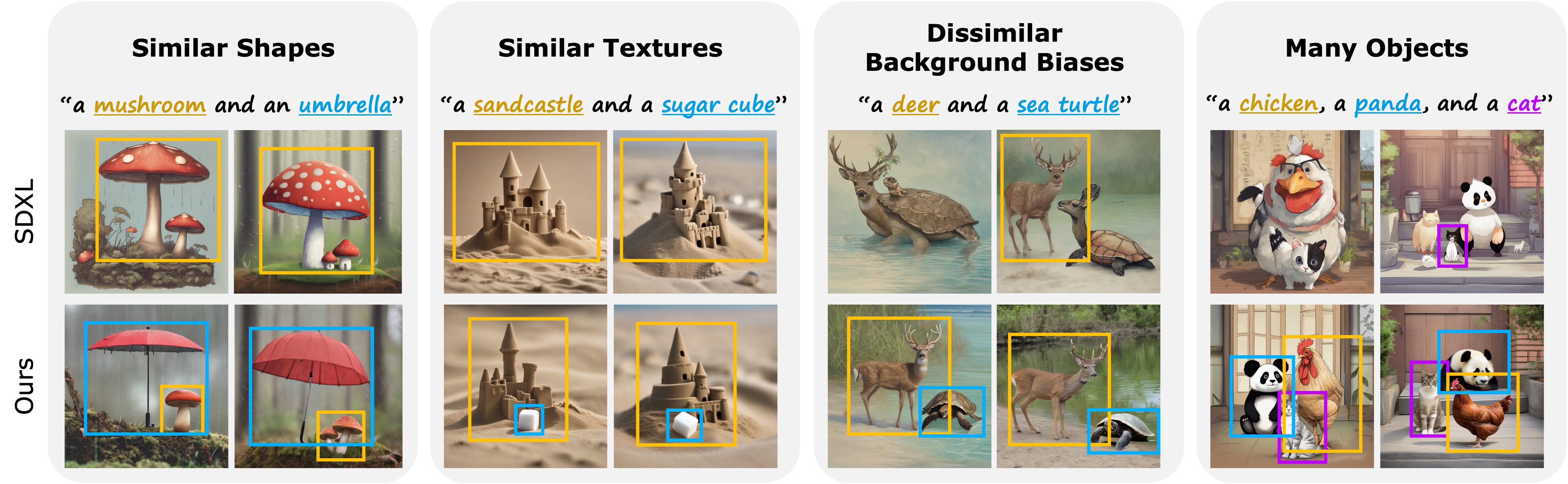}
  \captionof{figure}{Text-to-image generative models often struggle to generate images with multiple objects, especially in four scenarios where relationships between objects, referred to as \emph{inter-object relationships}, frequently lead to object neglect or object mixing: Similar Shapes, Similar Textures, Dissimilar Background Biases, and Many Objects. Our method mitigates these failures by modifying text embeddings before feeding them into the text-to-image model.
  }
  \label{fig:introduction}
  \vspace{-2.5mm}
\end{figure*}

\begin{abstract}
Recent progress in text-to-image~(T2I) generative models has led to significant improvements in generating high-quality images aligned with text prompts. However, these models still struggle with prompts involving multiple objects, often resulting in \emph{object neglect} or \emph{object mixing}. Through extensive studies, we identify four problematic scenarios, Similar Shapes, Similar Textures, Dissimilar Background Biases, and Many Objects, where inter-object relationships frequently lead to such failures. Motivated by two key observations about CLIP embeddings, we propose DOS~(Directional Object Separation), a method that modifies three types of CLIP text embeddings before passing them into text-to-image models. Experimental results show that DOS consistently improves the success rate of multi-object image generation and reduces object mixing. In human evaluations, DOS significantly outperforms four competing methods, receiving 26.24\%-43.04\% more votes across four benchmarks. These results highlight DOS as a practical and effective solution for improving multi-object image generation.
\end{abstract}

\begin{links}
    \link{Code}{https://github.com/dongnami/DOS}
\end{links}

\section{Introduction}
\label{sec:introduction}

Recent progress in text-to-image~(T2I) generative models has led to substantial improvements in generating high-quality images that closely align with text prompts~\cite{PodellEtAl2023, stabilityai2024stable35, blackforestlabs}. These models leverage text embeddings from multiple CLIP~\cite{radford2021learning} encoders, and optionally the T5~\cite{raffel2020exploring} encoder, to enhance image-text alignment. However, they still struggle to accurately generate images containing multiple objects. In this work, we identify four scenarios where this issue becomes particularly severe and demonstrate that modifying the CLIP text embeddings, without altering the image generation process within the text-to-image models, can substantially mitigate the problem.

Previous work on multi-object image generation has primarily focused on the relationships between nouns and their associated attributes~\cite{rassin2023linguistic, zhuang2024magnet, hu2024token}, often overlooking challenges that arise from interactions between multiple noun entities. However, through extensive experiments with state-of-the-art T2I models, we find that failures, specifically \emph{object neglect} or \emph{object mixing}, occur frequently in scenarios involving inter-object relationships, even when no attributes are present. To investigate these issues more thoroughly, we focus on prompts that primarily contain objects and avoid descriptive attributes. Following extensive exploration and experimentation, we identify four types of problematic scenarios. As shown in Figure 1, the first three categories represent specific types of inter-object relationships where failures become particularly severe: \emph{Similar Shapes}, \emph{Similar Textures}, and \emph{Dissimilar Background Biases}. \emph{Background bias} refers to the natural background typically associated with each object (e.g., `land' for `deer' vs. `sea' for `sea turtle'). The fourth, \emph{Many Objects}, captures the compounded effect of numerous inter-object relationships, which may include but are not limited to the first three types.

The motivation for our method stems from two previously reported observations regarding CLIP text embeddings used in T2I models: (1) CLIP employs a causal masking mechanism, where information from earlier tokens is mixed into later token embeddings, leading to \emph{information mix-ups} in the embeddings of later tokens~\cite{zhuang2024magnet, chen2024cat}, (2) the difference between two CLIP text embeddings appears to encode directional information, which can be indirectly observed through the behavior of T2I models~\cite{hu2024token}. Our method aims to mitigate information mix-ups in CLIP text embeddings by constructing directional vectors, referred to as \emph{separation vectors}, derived from the difference between two CLIP embeddings. These vectors are added to the original text embeddings to inject directional information that encourages the separation of object-specific information.

Our method, called DOS~(Directional Object Separation), constructs a separation vector for each object pair and each type of CLIP text embedding, namely semantic token embeddings of all object nouns, EOT embeddings, and pooled embeddings, to encode directional information that promotes separation. For each embedding type, the separation vectors of all object pairs are combined through a weighted average, where the weights reflect the difficulty of separating each corresponding pair. This weighted combination forms the DOS vector, which is then added to the original text embeddings.

Across four benchmarks corresponding to the problematic scenarios of \emph{Similar Shapes}, \emph{Similar Textures}, \emph{Dissimilar Background Biases}, and \emph{Many Objects}, DOS improves the success rate of multi-object image generation while reducing object mixing. In human evaluations, compared to four other models including the baseline, DOS received 26.24\%-43.04\% more votes than the second-best results across the four benchmarks. Since DOS does not alter the image generation process within the text-to-image model, its inference is approximately four times faster than well-known latent modification methods such as Attend-and-Excite~\cite{chefer2023attend}. These results demonstrate that DOS is both an effective and efficient algorithm for addressing object neglect and object mixing, which frequently arise from inter-object relationships in multi-object image generation.

\section{Related Work}
\label{sec:related_work}
Multi-object image generation involves generating images containing multiple objects, optionally along with their associated attributes. However, during generation, interactions among multiple objects, and optionally their associated attributes, often lead to failures such as object neglect, object mixing, or inaccurate attribute binding. A substantial body of work has been proposed to address these issues~\cite{chefer2023attend, hu2024ella, jiang2024comat}. Among them, Attend-and-Excite~\cite{chefer2023attend} demonstrates that iteratively updating the intermediate image latent can reduce such failures. This inspired a series of follow-up methods based on latent modification, such as CONFORM~\cite{meral2024conform}. While effective to some extent, these methods typically require expensive iterative gradient updates during inference and often introduce visual artifacts due to deviations from the latent distribution the model was trained on. To overcome these limitations, more recent approaches have focused on modifying only the text embeddings--without altering the intermediate image latent~\cite{chen2024cat, hu2024token, zhuang2024magnet}. Among these, TEBOpt~\cite{chen2024cat} proposes modifying the semantic token embeddings corresponding to each object to better separate them. This approach significantly reduces information bias~(i.e., cases where certain objects are more frequently generated than others), but shows only limited improvements in addressing object neglect and object mixing. In contrast, our method leverages \emph{directional information} to modify all types of CLIP text embeddings--not only semantic token embeddings, but also EOT and pooled embeddings--in a way that promotes object separation, significantly reducing both object neglect and object mixing in multi-object image generation.

\begin{figure*}[!t]
  \centering
  \includegraphics[width=0.92\textwidth]{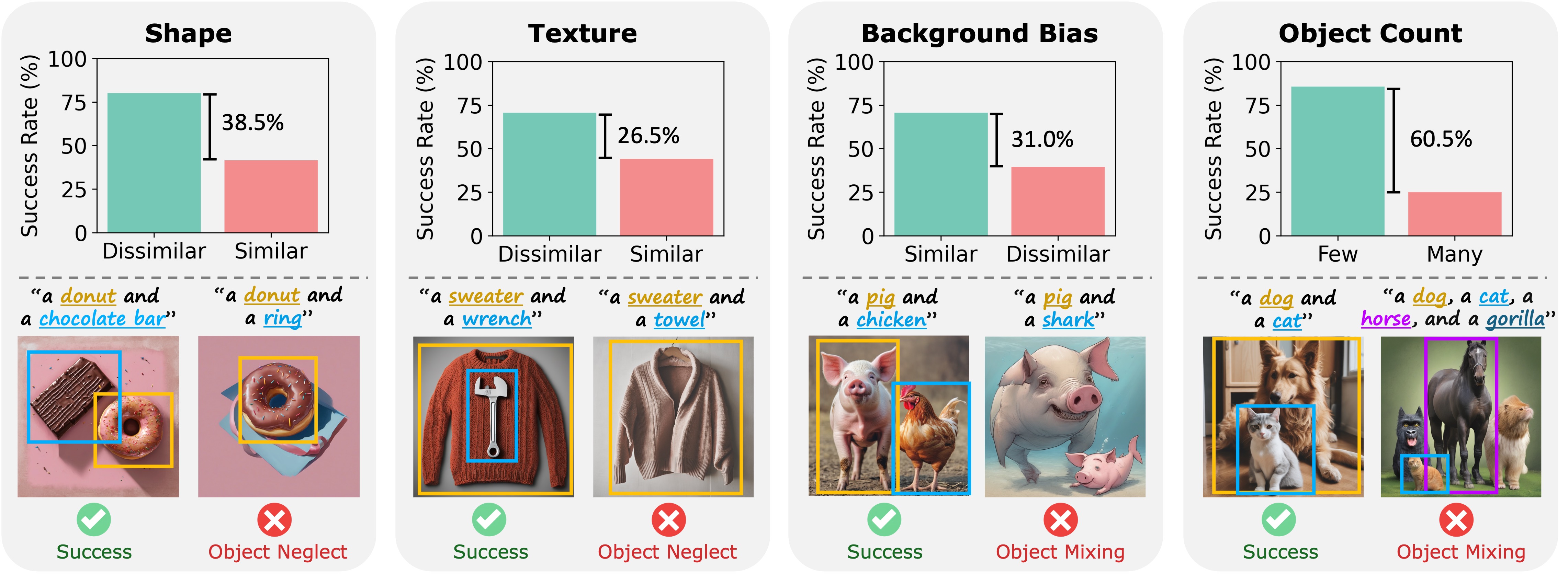}
  \captionof{figure}{Analysis of four aspects of inter-object relationships in multi-object image generation. The top row shows quantitative comparisons between two contrasting conditions for each aspect, while the bottom row shows a qualitative example for each condition. A lower success rate indicates more frequent failures, such as object neglect or object mixing.
  }
  \label{fig:preliminary_cases}
  \vspace{-2.5mm}
\end{figure*}

\section{Preliminary Analysis}
\label{sec:analysis}

In Figure~\ref{fig:introduction}, we identify four types of problematic scenarios involving inter-object relationships: \emph{Similar Shapes}, \emph{Similar Textures}, \emph{Dissimilar Background Biases}, and \emph{Many Objects}. To examine how each of these scenarios intensifies object neglect or object mixing, we analyze four corresponding aspects in multi-object image generation. For each aspect, we construct two sets of multi-object prompts that contrast in condition. For example, under the shape aspect, the \emph{dissimilar shapes} condition consists of 20 prompts featuring two objects with dissimilar shapes, following the template ``a/an \{$object~A$\} and a/an \{$object~B$\}.'' In contrast, the \emph{similar shapes} condition includes 20 prompts with two similarly shaped objects. The conditions for the other three aspects are defined analogously. Full details on the construction of all prompt sets are provided in Table~\ref{tab:appendix_condition_group_full_list_all} of Appendix~\ref{sec:appendix_four_condition_groupings}.

We generate images for each prompt using 10 random seeds, resulting in 200 images per condition. We then compute the success rate, defined as the percentage of images that exhibit neither object neglect nor object mixing. Figure~\ref{fig:preliminary_cases} presents both the quantitative results~(top row) and a qualitative example~(bottom row) for each condition across the four aspects. The observed differences in success rates between the two contrasting conditions range from 26.5\% to 60.5\%. These findings demonstrate that the four identified scenarios significantly intensify generation failures and motivate our focus on addressing them to reduce object neglect and mixing in multi-object image generation.

\section{DOS: Directional Object Separation in Text Embeddings}
\label{sec:method}

\paragraph{Background on CLIP text embeddings in text-to-image generative models.} Let $P$ be the input prompt for image generation, which is passed through CLIP text encoders to produce the CLIP text embedding: {$ [\, \bm{{c}}_{\mathrm{SOT}}, \bm{c}_{t_{1}}, \dots, \bm{c}_{t_{L}}, \bm{c}_{\mathrm{EOT}}, \bm{c}_{\mathrm{PAD}}, \dots, \bm{c}_{\mathrm{PAD}} \,]\in\mathbb{R}^{77 \times d}$}, where SOT, EOT, and PAD denote special tokens indicating the start-of-text, end-of-text, and padding respectively; $t_i$ denotes the $i$-th semantic token; and $d$ is the embedding dimension. Recent T2I models such as SDXL also use a pooled embedding {$\bm{c}_{\mathrm{pool}}$}, which is computed by applying a linear projection to the end-of-text~(EOT) embedding {$\bm{c}_{\mathrm{EOT}}$}. All these text embeddings are then used to condition the T2I model during image generation. Our method modifies these embeddings before passing them to the T2I model to mitigate object neglect and object mixing in multi-object image generation.

\begin{figure}[!t]
  \centering
  \includegraphics[width=\columnwidth]{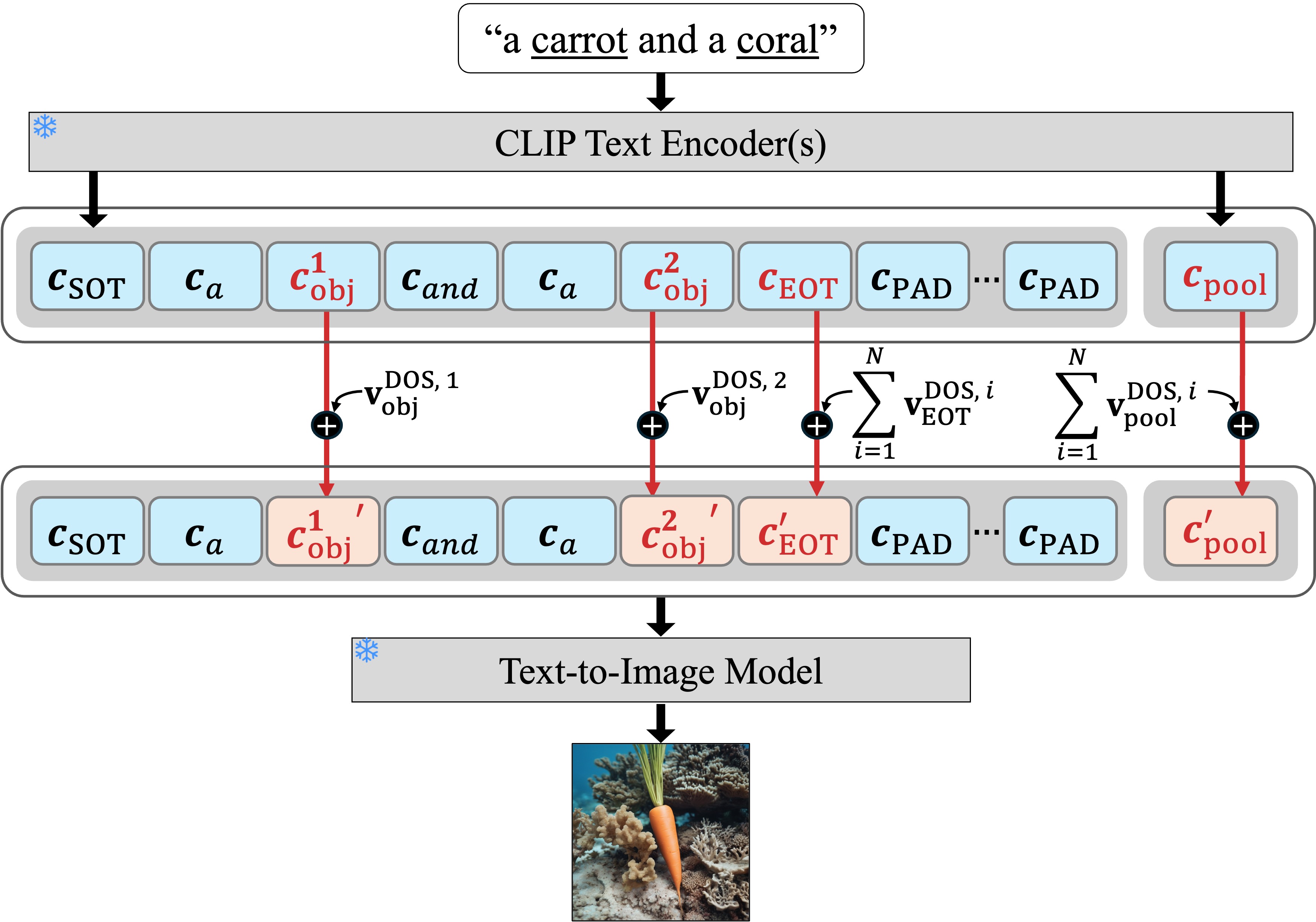}
  \captionof{figure}{Method Overview. DOS vectors are computed for the semantic token embeddings corresponding to object nouns, the EOT embedding, and the pooled embedding. These vectors are added to the corresponding CLIP text embeddings, namely $\bm{c}_{\mathrm{obj}}^{n}$ for $n=1,\dots,N$, $\bm{c}_{\mathrm{EOT}}$, and $\bm{c}_{\mathrm{pool}}$, before being passed into the text-to-image model. $N$ denotes the number of object nouns in the prompt.}
  \vspace{-3.5mm}
  \label{fig:method}
\end{figure}

\paragraph{Overview of the proposed method.} 
Suppose the input text prompt $P$ contains $N$ object nouns, where the $n$-th object noun is denoted by $obj_n$ for all $n \in \{1,2, \dots N \}$. Let $\bm{c}_{\mathrm{obj}}^{n}$ denote the semantic token embedding corresponding to $obj_n$. Additionally, let $\bm{c}_{\mathrm{EOT}}$ and $\bm{c}_{\mathrm{pool}}$ denote the EOT and pooled embeddings, respectively. Our method modifies all of these CLIP text embeddings by adding DOS vectors to each embedding. These DOS vectors are computed by aggregating separation vectors across all object pairs, where each separation vector defines a direction that distinguishes one object from the other in the pair. The following subsections describe: (1) how the separation vectors are computed for each object pair, (2) how the strength of each separation vector is adjusted prior to aggregation, and (3) how the adjusted vectors are combined to construct the final DOS vectors for each type of embedding. An overview of our method is illustrated in Figure~\ref{fig:method}.

\paragraph{Computing pairwise separation vectors.} 
The separation vectors for each object pair {$(obj_n, obj_m)$} are designed to encode directional information that promotes separation from $obj_m$ to $obj_n$. These vectors are computed differently depending on the type of text embedding. For the semantic token embedding {$\bm{c}_{\mathrm{obj}}^{n}$}, the separation vector is defined as the difference between two object-specific embeddings, {$\bm{c}_{\mathrm{obj}}^{\mathrm{pure},n}$} and {$\bm{c}_{\mathrm{obj}}^{\mathrm{pure},m}$}, each obtained from a pure prompt of the form $P_{\mathrm{pure}, i} = $ {``a \{$obj_i$\}''} for $i = n, m$, respectively:
\begin{equation}
{\mathbf{s}_{\mathrm{obj}}^{(n,m)} = \bm{c}_{\mathrm{obj}}^{\mathrm{pure},n} - \bm{c}_{\mathrm{obj}}^{\mathrm{pure},m}} 
\end{equation}
Using pure prompts ensures that {$\bm{c}_{\mathrm{obj}}^{\mathrm{pure},n}$} and {$\bm{c}_{\mathrm{obj}}^{\mathrm{pure},m}$} capture clean, object-specific representations by avoiding undesirable information mix-ups across tokens.
In contrast, for EOT and pooled embeddings, we intentionally exploit information mix-ups to define separation vectors. Specifically, we construct two contrastive prompts: {$P_{\mathrm{sep}, (n,m)} = $ ``a \{$obj_n$\} separated from a \{$obj_m$\}''} and {$P_{\mathrm{mix}, (n,m)} = $ ``a \{$obj_n$\} mixed with a \{$obj_m$\}''}. From these, we extract the corresponding EOT and pooled embeddings, denoted as {$\bm{c}_{\mathrm{EOT/pool}}^{\mathrm{sep}, (n,m)}$} and {$\bm{c}_{\mathrm{EOT/pool}}^{\mathrm{mix}, (n,m)}$}, and define the separation vectors as:
\begin{equation}
{\mathbf{s}_{\mathrm{EOT/pool}}^{(n,m)} = \bm{c}_{\mathrm{EOT/pool}}^{\mathrm{sep}, (n,m)} - \bm{c}_{\mathrm{EOT/pool}}^{\mathrm{mix}, (n,m)}}
\end{equation}
While we use the unified notation {$\mathbf{s}_{\mathrm{EOT/pool}}^{(n,m)}$} for brevity, these separation vectors are computed independently for the EOT and pooled embeddings. 
Overall, our separation vector design extracts directional signals that promote object separation: by avoiding information mix-ups for semantic token embeddings, and by deliberately exploiting contrastive prompts to leverage such mix-ups for EOT and pooled embeddings.

\paragraph{Computing pairwise adaptive strengths.} In Figure~\ref{fig:preliminary_cases}, we observed that certain object pairs are more susceptible to object neglect or object mixing than others. Motivated by this observation, we adjust the strength of the separation vectors for each object pair based on their visual~(shape and texture) similarities and background bias dissimilarities. To this end, we first select 42 representative words covering a diverse range of object shapes and textures through interactions with GPT‑o4‑mini‑high~\cite{openai_gpt-o4-mini-high_2025} and define prompts of the form {$P_{\mathrm{attr}, (k,n)} = $ ``a \{$attribute_k$\} \{$obj_n$\}’’} for all $k \in \{ 1, 2, \dots, 42 \}$. Similarly, we select 36 representative background phrases and define prompts of the form {$P_{\mathrm{bg}, (l,n)} = $``\{$background_l$\}, there is a \{$obj_n$\}’’} for all $l \in \{ 1, 2, \dots, 36 \}$. We then compare the text embeddings {$\bm{c}_{\mathrm{obj/EOT/pool}}^{\mathrm{pure}, n}$}, obtained from the pure prompt $P_{\mathrm{pure}, n} = $ {``a \{$obj_n$\}''}, with embeddings {$\bm{c}_{\mathrm{obj/EOT/pool}}^{\mathrm{attr}, (k,n)}$} and {$\bm{c}_{\mathrm{obj/EOT/pool}}^{\mathrm{bg}, (l,n)}$} obtained from the attribute and background prompts, respectively. Cosine similarity is used for this comparison:
\begin{align}
\mathbf{a}_{\tau}^{\mathrm{attr}, n}[k] &= \text{sim} (\bm{c}_{\tau}^{\mathrm{pure}, n}, \bm{c}_{\tau}^{\mathrm{attr}, (k, n)}), \\
\mathbf{a}_{\tau}^{\mathrm{bg}, n}[l] &= \text{sim} (\bm{c}_{\tau}^{\mathrm{pure}, n}, \bm{c}_{\tau}^{\mathrm{bg}, (l, n)}),
\end{align}
for $k \in \{ 1, 2, \dots, 42 \}$, $l \in \{ 1, 2, \dots,36 \}$, and $\tau \in \{ \mathrm{obj, EOT, pool}\}$, where $\tau$ denotes the type of text embedding. The resulting lists of cosine similarities $\mathbf{a}_{\tau}^{\mathrm{attr}, n}$ and $\mathbf{a}_{\tau}^{\mathrm{bg}, n}$ reflect how closely $obj_n$ aligns with the predefined visual attributes and background contexts. We compute the same lists for $obj_m$, and define the adaptive strength for the object pair ($obj_n$, $obj_m$) as:
\begin{align}
\alpha_{\tau}^{(n,m)} = \max \Bigl\{ 
&\sigma \bigl( \rho( \mathbf{a}_{\tau}^{\mathrm{attr},n}, \mathbf{a}_{\tau}^{\mathrm{attr},m} ); x_{\tau, 0}^{\mathrm{attr}} \bigr), \notag \\
&\sigma \bigl( 1 - \rho( \mathbf{a}_{\tau}^{\mathrm{bg},n}, \mathbf{a}_{\tau}^{\mathrm{bg},m} ); x_{\tau, 0}^{\mathrm{bg}} \bigr) \Bigr\},
\label{eqn:adaptive strength}
\end{align}
for $\tau \in \{ \mathrm{obj, EOT, pool}\}$, where $\rho (\cdot, \cdot)$ denotes Pearson correlation, which is used to avoid mean and scaling effects, and $\sigma(x;x_{\tau, 0})$ is a shifted tempered sigmoid function defined as:
\begin{align}
\rho(x,y) &= \frac{\sum_t (x_t - \bar x)(y_t - \bar y)} {\sqrt{\sum_t (x_t - \bar x)^2} \;\sqrt{\sum_t (y_t - \bar y)^2}},\\[0.5em]
\label{eqn:tempered sigmoid function}
\sigma(x; x_{\tau, 0}) &= \bigl(1 + \exp\{-(x - x_{\tau, 0})/T\}\bigr)^{-1}
\end{align}
Here, the offset $x_{\tau, 0}^{\mathrm{attr}}$ is the mean of {$\rho( \mathbf{a}_{\tau}^{\mathrm{attr},i}, \mathbf{a}_{\tau}^{\mathrm{attr},j} )$} over all object pairs {$(obj_i, obj_j)$} from the MS-COCO~\cite{lin2014microsoft} dataset. Similarly, $x_{\tau, 0}^{\mathrm{bg}}$ is the mean of {$\bigl( 1 - \rho( \mathbf{a}_{\tau}^{\mathrm{bg},i}, \mathbf{a}_{\tau}^{\mathrm{bg},j}) \bigr)$} over the same object pairs. These adaptive strengths promote stronger separation for object pairs that are more likely to be mixed or neglected, thereby helping to reduce object neglect and mixing in multi-object image generation. In Appendix~\ref{sec:appendix_implementation_details}, we provide the full list of 42 representative words describing object shapes and textures and 36 representative background phrases, which are used to construct the prompts $P_{\mathrm{attr}, (k,n)}$ and $P_{\mathrm{bg}, (l,n)}$, respectively.

\paragraph{Updating text embeddings.} Finally, the DOS vectors for the semantic token, EOT, and pooled embeddings are computed as a weighted average of the separation vectors, where each weight corresponds to the adaptive strength of a given object pair:
\begin{equation}
\mathbf{v}_{\tau}^{\mathrm{DOS}, n} = \frac{1}{N-1} \sum_{m\neq n} \alpha_{\tau}^{(n,m)}\,\mathbf{s}_{\tau}^{(n,m)},
\end{equation}
where $\tau \in \{ \mathrm{obj, EOT, pool}\}$, and $n \in \{ 1, 2, \dots, N \}$. These DOS vectors are then added to the original text embeddings, which are subsequently passed into the T2I model. Specifically, semantic token embeddings corresponding to object nouns are updated as:
\begin{equation}
\label{eqn:DOS_update_semantic token embeddings}
{\bm{c}_{\mathrm{obj}}^n}' = \bm{c}_{\mathrm{obj}}^n + \mathbf{v}_{\mathrm{obj}}^{\mathrm{DOS}, n},
\end{equation}
for $n \in \{ 1, 2, \dots, N \}$. If {$obj_n$} consists of multiple tokens, their semantic token embeddings are first averaged before computing the DOS vector, and the resulting DOS vector is then added to each of the semantic token embeddings that constitute $obj_n$. In addition, the EOT and pooled embeddings are updated by summing the DOS vectors across all objects:
\begin{align}
\label{eqn:DOS_update_semantic EOT/pooled embeddings}
\bm{c}_{\mathrm{EOT/pool}}' &= \bm{c}_{\mathrm{EOT/pool}} + \sum_{i=1}^N \mathbf{v}_{\mathrm{EOT/pool}}^{\mathrm{DOS}, i}
\end{align}
While we use the unified notation {$\bm{c}_{\mathrm{EOT/pool}}$} and {$\bm{c}_{\mathrm{EOT/pool}}’$} for brevity, the EOT and pooled embeddings are computed and updated independently. This aggregation across all objects is intended to mitigate the compounded effects of inter-object relationships in multi-object image generation.

\begin{table*}[t]
  \renewcommand{\arraystretch}{0.85}
  \setlength{\extrarowheight}{0pt}

  \centering
  \setlength{\tabcolsep}{5.5pt}
  \scriptsize
  \resizebox{0.82\textwidth}{!}{%
    \begin{tabular}{@{}llcc cc cc cc@{}}
      \toprule
      \multirow{2}{*}{Base Model} & \multirow{2}{*}{Method}
        & \multicolumn{2}{c}{Similar Shapes}
        & \multicolumn{2}{c}{Similar Textures}
        & \multicolumn{2}{c}{Dissimilar Background Biases}
        & \multicolumn{2}{c}{Many Objects} \\
      \cmidrule(lr){3-4} \cmidrule(lr){5-6} \cmidrule(lr){7-8} \cmidrule(lr){9-10}
      & 
        & SR↑ & MR↓
        & SR↑ & MR↓
        & SR↑ & MR↓
        & SR↑ & MR↓ \\
      \midrule
      \multirow{5}{*}{SDXL}
        & Base    & 48.00\% & 6.50\%  & 58.00\% & 7.50\%  & 46.00\% & 22.50\% & 23.00\% & 27.50\% \\
        & TEBOpt  & 52.00\% & 5.00\%  & 61.00\% & 4.00\%  & 44.50\% & 23.00\% & 24.00\% & 27.00\% \\
        & A\&E    & 60.50\% & 6.00\%  & 67.50\% & 5.00\%  & 53.50\% & 25.00\% & 28.50\% & 28.50\% \\
        & CONFORM & 54.00\% & 6.50\%  & 64.50\% & 9.00\%  & 55.50\% & 22.00\% & 37.50\% & 26.50\% \\
        & Ours    & \textbf{64.00\%} & \textbf{3.50\%}
                  & \textbf{71.50\%} & \textbf{3.50\%}
                  & \textbf{68.50\%} & \textbf{17.00\%}
                  & \textbf{48.00\%} & \textbf{15.50\%} \\
      \midrule
      \multirow{2}{*}{SD3.5}
        & Base    & 75.50\% & 4.00\%  & 79.00\% & 6.00\%  & 78.00\% & 17.50\% & 70.00\% & 16.50\% \\
        & Ours    & \textbf{81.00\%} & \textbf{3.00\%}
                  & \textbf{87.50\%} & \textbf{3.00\%}
                  & \textbf{85.50\%} & \textbf{13.50\%}
                  & \textbf{76.50\%} & \textbf{10.50\%} \\
      \bottomrule
    \end{tabular}%
  }
  \caption{Quantitative comparison across four benchmarks: Similar Shapes, Similar Textures, Dissimilar Background Biases, and Many Objects.
SR denotes Success Rate~(↑ higher is better), and MR denotes Mixture Rate~(↓ lower is better).}
  \label{tab:performance_comparison}
  \vspace{-3.5mm}
  \renewcommand{\arraystretch}{1.0}
\end{table*}

\begin{table}[t]
  \renewcommand{\arraystretch}{0.85}
  \setlength{\extrarowheight}{0pt}

  \centering
  \setlength{\tabcolsep}{3pt}
  \scriptsize
  \resizebox{0.9\columnwidth}{!}{%
    \begin{tabular}{@{}lcccc@{}}
      \toprule
      Method
        & \makecell{Similar\\[-0.8ex]Shapes}
        & \makecell{Similar\\[-0.8ex]Textures}
        & \makecell{Dissimilar\\[-0.8ex]Background Biases}
        & \makecell{Many\\[-0.8ex]Objects} \\
      \midrule
      SDXL        & \hphantom{0}7.59\%  & \hphantom{0}9.38\%  & \hphantom{0}3.59\%  & \hphantom{0}3.21\%  \\
      TEBOpt      & \hphantom{0}7.24\%  & \hphantom{0}7.03\%  & \hphantom{0}4.22\%  & \hphantom{0}4.82\%  \\
      A\&E        & 13.62\% & 14.84\% & 22.81\% & 11.43\% \\
      CONFORM     & 12.76\% & 16.88\% & 12.19\% & 12.14\% \\
      Ours        & \textbf{50.52\%} & \textbf{43.12\%} & \textbf{53.44\%} & \textbf{55.18\%} \\
      No winner   & \hphantom{0}8.28\%  & \hphantom{0}8.75\%  & \hphantom{0}3.75\%  & 13.21\% \\
      \bottomrule
    \end{tabular}%
  }
  \caption{Results of human preference studies across four benchmarks. All methods are based on SDXL. 
  }
  \label{tab:human-sr}
  \renewcommand{\arraystretch}{1.0}
\end{table}

\begin{table}[t]
  \centering
  \setlength{\tabcolsep}{3.5pt}
  \scriptsize
  \resizebox{0.9\columnwidth}{!}{%
    \begin{tabular}{@{}lccccc@{}}
      \toprule
      & SDXL & TEBOpt & A\&E & CONFORM & Ours \\
      \midrule
      Inference Time (s)
        & 12.97 & 13.50 & 58.83 & 58.48 & 13.87 \\
      \bottomrule
    \end{tabular}%
  }
  \caption{Inference time comparison~(in seconds). All values are averaged over 10 runs. All methods are based on SDXL.}
  \label{tab:comp-cost-horizontal}
  \vspace{-3.5mm}
\end{table}

\begin{figure*}[t]
  \centering
  \includegraphics[width=0.91\textwidth]{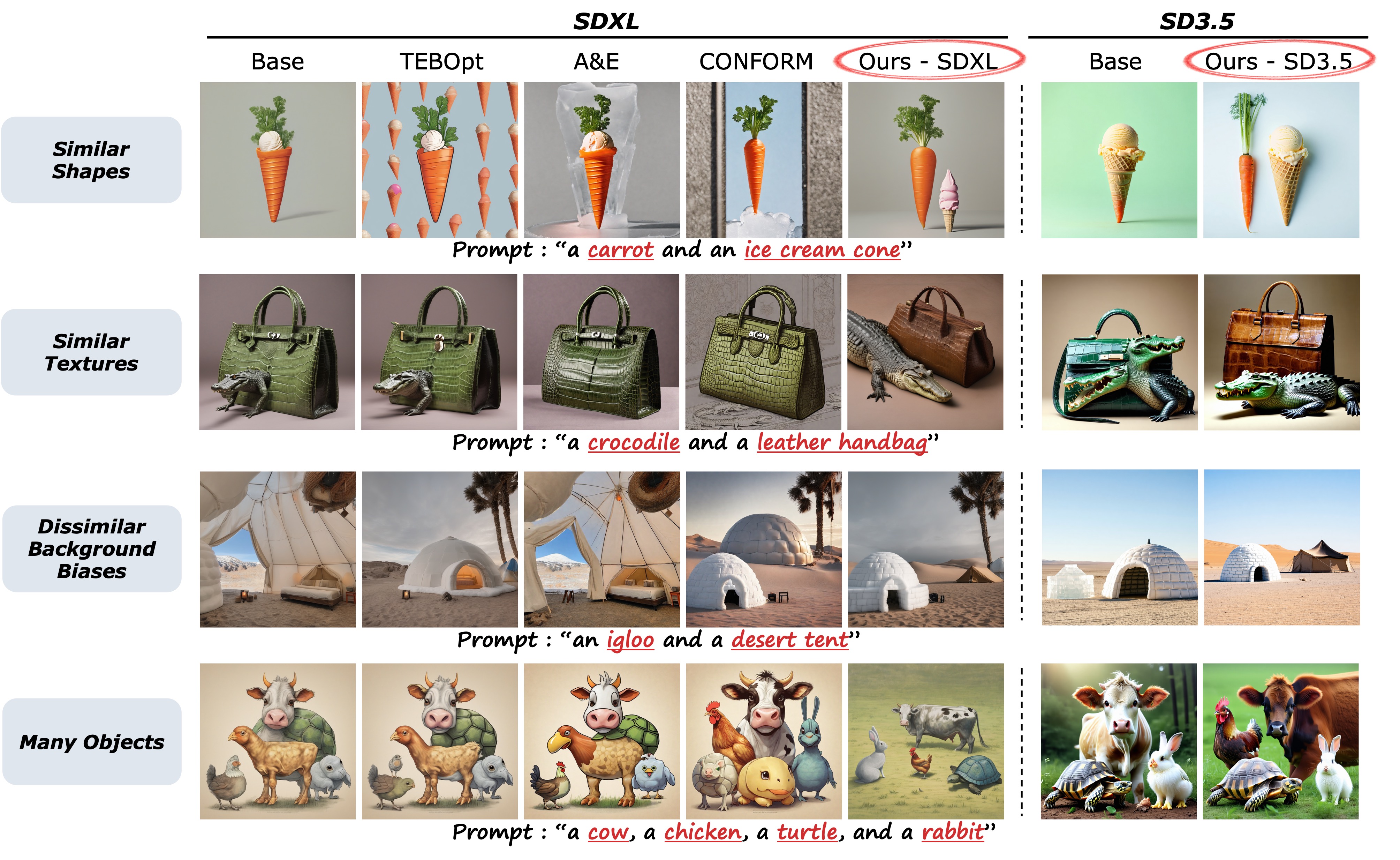}
  \captionof{figure}{Qualitative comparison across four benchmarks: \emph{Similar Shapes}, \emph{Similar Textures}, \emph{Dissimilar Background Biases}, and \emph{Many Objects}. More results are shown in Figures~\ref{fig:qualitative_extension_1}-\ref{fig:qualitative_extension_3} of Appendix~\ref{sec:appendix_additional_qualitative_results}.
  }
  \vspace{-3mm}
  \label{fig:qualitative_comparison}
\end{figure*}

\begin{figure*}[t]
  \centering
  \includegraphics[width=0.8\textwidth]{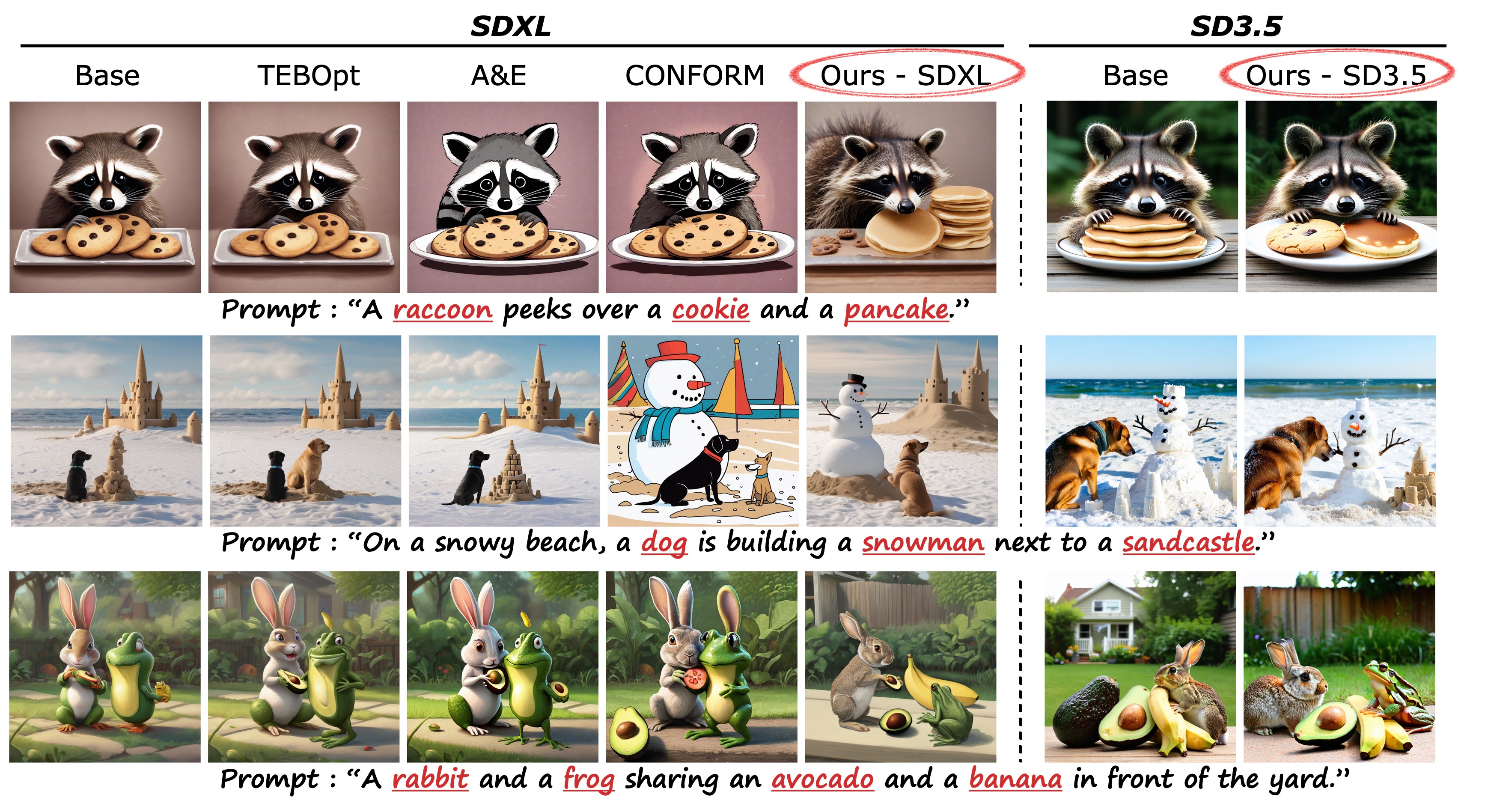}
  \captionof{figure}{Qualitative comparison on complex prompts. More results are shown in Figure~\ref{fig:general_prompt_extension} of Appendix~\ref{sec:appendix_additional_qualitative_results}.
  }
  \vspace{-2.5mm}
  \label{fig:general_prompt}
\end{figure*}

\section{Experiments}
\label{sec:experiment}

\subsection{Experimental Setup}

\paragraph{Implementation details.} Our method is based on SDXL~\cite{PodellEtAl2023} and has also been implemented on the recently released Stable Diffusion 3.5~(SD3.5)~\cite{stabilityai2024stable35}. We apply DOS updates to all CLIP text embeddings used by each model. For inference, we use a guidance scale of 5.0 with 50 denoising steps for SDXL, and a guidance scale of 7.0 with 28 steps for SD3.5, which are the official default configurations. The temperature $T$ of the shifted tempered sigmoid functions in Eq.~\ref{eqn:tempered sigmoid function} is set to $T=0.6$ for all base models, determined through a light validation process. Additional implementation details are provided in Appendix~\ref{sec:appendix_implementation_details}.

\paragraph{Dataset.} To focus on the four problematic scenarios in multi-object image generation, we construct four corresponding benchmarks: Similar Shapes, Similar Textures, Dissimilar Background Biases, and Many Objects. Each of the first three benchmarks consists of 50 two-object prompts, following the template: ``a/an $\{object~A\}$ and a/an $\{object~B\}$.'' The object pairs for these benchmarks are selected through interactions with GPT-o4-mini-high~\cite{openai_gpt-o4-mini-high_2025}. For the Many Objects benchmark, we extend the animal list used in Attend-and-Excite to include 17 animals, from which we randomly sample to create 25 three-object prompts and 25 four-object prompts. For each method, we generate 4 images per prompt using different random seeds, resulting in 200 generated images per benchmark. While the main experiment for Many Objects focuses on the animal list, we present results for an extended benchmark in Table~\ref{tab:appendix_many_objects_v2} of Appendix~\ref{sec:extend_object_categories} that also includes electronics, vehicles, and fruits. Detailed information about the construction of each benchmark is provided in Appendix~\ref{sec:appendix_benchmark_details}.

\paragraph{Evaluation metrics.}
We adjust the VLM-based metric from EnMMDiT~\cite{wei2024enhancing} to measure both the success rate~(SR) and mixture rate~(MR) for each generated image. SR evaluates whether all the objects specified in the prompt are clearly present in the image, while MR assesses whether any of the objects appear mixed. Specifically, we provided GPT-4o-mini with the generated image and the text prompt used for image generation, asking it to classify each object in the prompt as (1) fully intact, (2) mixed, or (3) absent. SR is calculated as the ratio of images where all objects are classified as fully intact, and MR is calculated as the ratio of images where one or more objects are classified as mixed. A high SR score and a low MR score indicate superior image-text alignment. We further validate the consistency of the results across two additional evaluators in Table~\ref{tab:appendix_re-evaluation} of Appendix~\ref{sec:appendix_validation_metrics}. Additional details about the VLM-based metrics are provided in Appendix~\ref{sec:appendix_evaluation_metrics}. 

\subsection{Experimental Results}
\label{subsec:Experimental Results}
We compare our method with TEBOpt~\cite{chen2024cat}, Attend-and-Excite (A\&E)~\cite{chefer2023attend}, and CONFORM~\cite{meral2024conform} using SDXL as the base model. For SD3.5, we only compare our method with the baseline model in our main experiment. Further comparisons on SD3.5 against recent methods, TEBOpt~\cite{chen2024cat} and Self-Cross~\cite{qiu2025self}, are provided in Table~\ref{tab:appendix_sd3-5-comparison} of Appendix~\ref{sec:appendix_comparison_on_sd3_5}.

\paragraph{Qualitative comparison.}
Qualitative comparisons for each problematic scenario are shown in Figure~\ref{fig:qualitative_comparison}. Across all scenarios and base models, our method successfully generates all objects specified in the prompts, while other methods often exhibit object neglect or object mixing. For example, given the prompt ``a carrot and an ice cream cone,'' our method successfully generates both objects without mixing, whereas the other methods fail to do so. In Figure~\ref{fig:general_prompt}, we also compare results on more complex prompts, where our method continues to show better generations. Additional qualitative results, including examples with complex prompts, are provided in Figures~\ref{fig:qualitative_attribute_benchmark_1}-\ref{fig:general_prompt_extension} of Appendix~\ref{sec:appendix_additional_qualitative_results}.

\paragraph{Quantitative comparison.}
Table~\ref{tab:performance_comparison} presents quantitative comparisons across all four benchmarks in multi-object image generation. Across all benchmarks and base models, our method achieves the highest success rate~(SR)~(higher is better) and the lowest mixture rate~(MR)~(lower is better). Specifically, our method outperforms TEBOpt, which is the other text embedding modification method, by 10.50\%-24.00\% in SR. It also surpasses the latent modification methods A\&E and CONFORM, which generally require much higher inference time compared to ours. On the recently released SD3.5 model, our method further improves over the baseline, demonstrating that its effectiveness extends to newer architectures.

\paragraph{Human preference study.} We conducted human preference studies across all four benchmarks in multi-object image generation. In each survey, participants were presented with five images generated from the same text prompt but using different SDXL-based methods, and asked to select the one that best represents all the objects specified in the prompt. Participants were instructed to avoid images with object neglect or object mixing. Using Amazon Mechanical Turk~(MTurk), we collected 580, 640, 640, 560 responses from 29, 32, 32, 28 valid participants for each survey, respectively. As shown in Table~\ref{tab:human-sr}, our method significantly outperforms the others, receiving 26.24\%-43.04\% more votes than the second-best results across the four benchmarks. Additional details on the human preference studies, including survey questions and participant filtering criteria, are provided in Appendix~\ref{sec:appendix_human_evaluation}.

\subsection{Analysis}
\paragraph{Inference time comparison.}

To demonstrate the time efficiency of our method, we compare the inference times of various SDXL-based methods. As shown in Table~\ref{tab:comp-cost-horizontal}, text embedding modification methods, including TEBOpt and our method, exhibit inference times similar to the baseline model, indicating minimal time overhead for modifying text embeddings. In contrast, latent modification methods A\&E and CONFORM require approximately 4-5 times more computational time than the baseline, due to the iterative gradient updates during image generation. This highlights that our method offers both an efficient and effective solution to the challenges of multi-object generation.

\paragraph{Ablation on text embedding types.}
Our method applies DOS updates to three types of CLIP text embeddings: semantic token embeddings corresponding to object nouns, EOT embeddings, and pooled embeddings. In Table~\ref{tab:separating_vectors}, we ablate the type of embeddings to which DOS is applied. The results show that applying DOS to either the semantic token embeddings or the EOT/pooled embeddings leads to notable improvements in SR, while MR is more significantly reduced when DOS is applied to EOT/pooled embeddings. This is likely because EOT and pooled embeddings tend to capture the overall semantics of the prompt~\cite{wu2024relation}, influencing the overall spatial structure of the generated images when modified. Finally, applying DOS to all three types of embeddings results in the best performance on both SR and MR. Qualitative examples for this ablation study are shown in Figure~\ref{fig:ablation_qualitative} of Appendix~\ref{sec:appendix_qualitative_ablation_on_text_embedding_types}.

\paragraph{Ablation on adaptive strengths.}
To examine the effect of the adaptive strengths defined in Eq.~\ref{eqn:adaptive strength}, we perform an ablation by setting all adaptive strengths to a fixed value of 0.5. The results, shown in Table~\ref{tab:strength-comparison}, indicate that using the fixed strength for all separation vectors already leads to substantial performance improvements compared to the baseline. This suggests that the modification directions specified by the separation vectors are effective on its own, even without adaptive strength modulation. However, applying adaptive strength results in even better performance, indicating that modulating the strength of separation differently for each object pair further enhances image generation with multiple objects. An additional ablation study is provided in Table~\ref{tab:appendix_sensitivity_adaptive_strength} of Appendix~\ref{sec:appendix_sensitivity_ablation}, examining sensitivity to predefined words and phrases used to compute adaptive strengths.

\begin{table}[t]
  \centering
  \small
  \setlength{\tabcolsep}{2pt}
  \resizebox{\columnwidth}{!}{%
    \begin{tabular}{@{}l cc cc cc cc@{}}
      \toprule
      Method
        & \multicolumn{2}{c}{\makecell{Similar\\Shapes}}
        & \multicolumn{2}{c}{\makecell{Similar\\Textures}}
        & \multicolumn{2}{c}{\makecell{Dissimilar\\Background Biases}}
        & \multicolumn{2}{c}{\makecell{Many\\Objects}} \\
      \cmidrule(lr){2-3} \cmidrule(lr){4-5} \cmidrule(lr){6-7} \cmidrule(lr){8-9}
      & SR↑ & MR↓
      & SR↑ & MR↓
      & SR↑ & MR↓
      & SR↑ & MR↓ \\
      \midrule
      SDXL
        & 48.0\% &  6.5\%
        & 58.0\% &  7.5\%
        & 46.0\% & 22.5\%
        & 23.0\% & 27.5\% \\
      +obj
        & 58.0\% &  5.0\%
        & 64.0\% &  6.5\%
        & 51.0\% & 23.0\%
        & 30.0\% & 29.5\% \\
      +EOT/pool
        & 57.0\% &  4.5\%
        & 68.0\% &  5.0\%
        & 59.5\% & 17.5\%
        & 36.5\% & 16.0\% \\
      +obj+EOT/pool
        & \textbf{64.0\%} & \textbf{3.5\%}
        & \textbf{71.5\%} & \textbf{3.5\%}
        & \textbf{68.5\%} & \textbf{17.0\%}
        & \textbf{48.0\%} & \textbf{15.5\%} \\
      \bottomrule
    \end{tabular}%
  }
  \caption{Ablation study on the types of text embeddings modified by DOS. We compare applying DOS to only the semantic token embeddings~(obj) or only the EOT/pooled embeddings, against our default approach that modifies all.}
  \label{tab:separating_vectors}
\end{table}

\begin{table}[t]
  \centering
  \small
  \setlength{\tabcolsep}{2pt}
  \resizebox{\columnwidth}{!}{%
    \begin{tabular}{@{}l cc cc cc cc@{}}
      \toprule
      Method
        & \multicolumn{2}{c}{\makecell{Similar\\Shapes}}
        & \multicolumn{2}{c}{\makecell{Similar\\Textures}}
        & \multicolumn{2}{c}{\makecell{Dissimilar\\Background Biases}}
        & \multicolumn{2}{c}{\makecell{Many\\Objects}} \\
      \cmidrule(lr){2-3} \cmidrule(lr){4-5} \cmidrule(lr){6-7} \cmidrule(lr){8-9}
      & SR↑ & MR↓
      & SR↑ & MR↓
      & SR↑ & MR↓
      & SR↑ & MR↓ \\
      \midrule
      SDXL
        & 48.0\% &  6.5\%
        & 58.0\% &  7.5\%
        & 46.0\% & 22.5\%
        & 23.0\% & 27.5\% \\
      Ours ($\alpha=0.5$)
        & 60.5\% &  4.5\%
        & 67.5\% &  6.5\%
        & 67.0\% & 19.5\%
        & 41.0\% & 25.5\% \\
      Ours (adaptive $\alpha$'s)
        & \textbf{64.0\%} & \textbf{3.5\%}
        & \textbf{71.5\%} & \textbf{3.5\%}
        & \textbf{68.5\%} & \textbf{17.0\%}
        & \textbf{48.0\%} & \textbf{15.5\%} \\
      \bottomrule
    \end{tabular}%
  }
  \caption{Ablation study comparing adaptive strengths against a fixed strength of $\alpha=0.5$.}
  \label{tab:strength-comparison}
  \vspace{-3mm}
\end{table}

\begin{figure}[t]
  \centering
  \vspace{-2.5mm}
  \includegraphics[width=0.8\columnwidth]{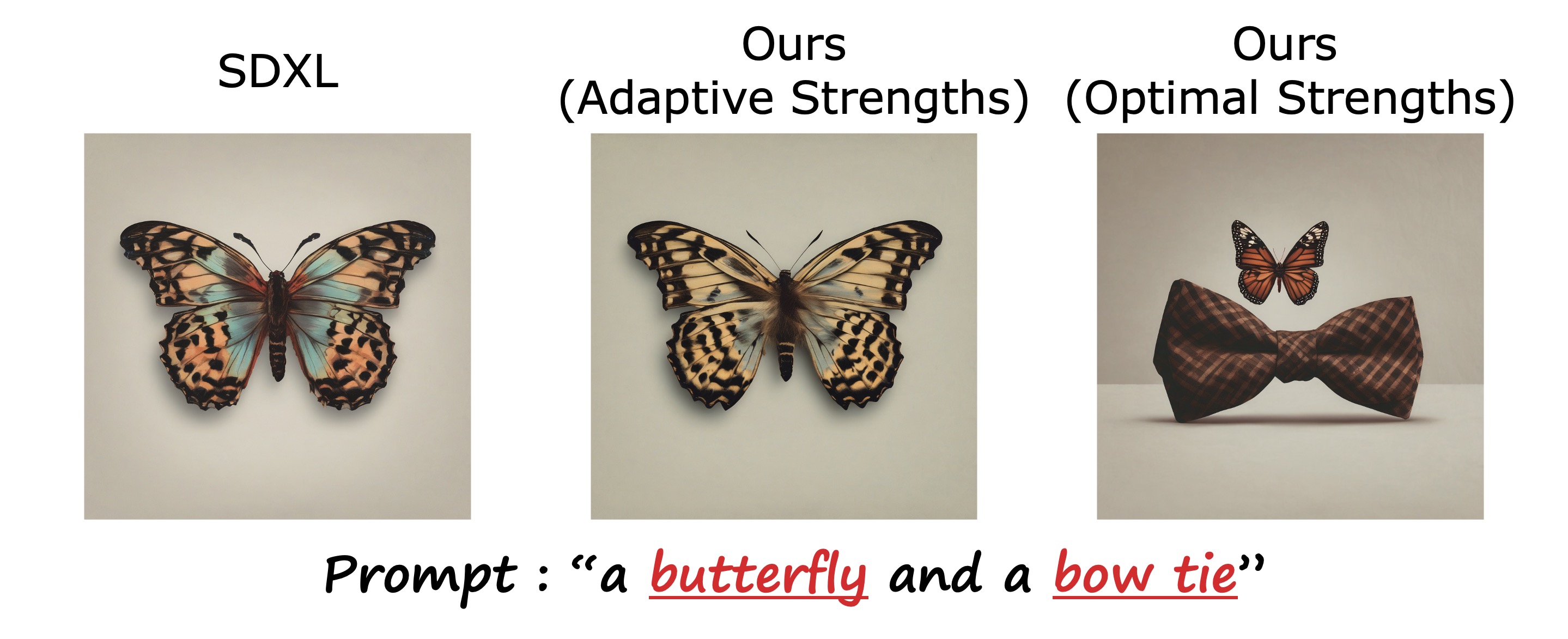}
  \caption{Example where adaptive strengths fail to properly scale separation vectors, leading to object neglect.}
  \vspace{-3.5mm}
  \label{fig:limitation}
\end{figure}

\section{Limitation}
\label{sec:limitation}
We demonstrated that the adaptive strengths~(defined in Eq.~\ref{eqn:adaptive strength}) in DOS effectively adjust the magnitude of separation vectors, leading to improved multi-object image generation~(see Table~\ref{tab:strength-comparison}). However, as shown in Figure~\ref{fig:limitation}, it does not always yield the optimal strengths. In this figure, we compare the default adaptive strength with the optimal strength obtained via grid search over 121 hyperparameter settings (an 11$\times$11 grid with a 0.1 interval). Although such cases are rare, they suggest that there is still room to improve how the strength of each separation vector is determined, potentially by incorporating additional criteria beyond shape, texture, and background bias.

\section{Conclusion}
\label{sec:conclusion}
In this work, we propose DOS (Directional Object Separation), an efficient and effective method for modifying text embeddings to improve multi-object image generation. Through extensive studies, we identify four problematic scenarios where inter-object relationships frequently cause failures such as object neglect or object mixing. Motivated by two key observations about CLIP embeddings, we construct DOS vectors and add them to three types of CLIP text embeddings before passing them into T2I models. This approach not only offers a practical solution to challenges in multi-object image generation but also demonstrates how the properties of CLIP embeddings can be leveraged to address limitations in downstream models that rely on them.

\section*{Acknowledgments}

This work was supported by Basic Science Research Program through the National Research Foundation of Korea (NRF) funded by the Ministry of Education (No. RS-2022-NR070876) and in part by Institute of Information \& communications Technology Planning \& Evaluation (IITP) grant funded by the Korea government (MSIT) ([NO.RS-2021-II211343, Artificial Intelligence Graduate School Program (Seoul National University)], [No.RS-2023-00235293, Development of autonomous driving big data processing, management, search, and sharing interface technology to provide autonomous driving data according to the purpose of usage]).

\bibliography{aaai2026} 


\clearpage
\newpage
\appendix

\setcounter{secnumdepth}{2}

\twocolumn[

{\centering
  {\LARGE\bfseries DOS: Directional Object Separation in Text Embeddings\\
    for Multi-Object Image Generation\par}
  \vspace{1ex}
  {\large\itshape Supplementary Material\par}
  \vspace{1ex}}




\section{Details on Preliminary Analysis}
\label{sec:appendix_four_condition_groupings}
In Table~\ref{tab:appendix_condition_group_full_list_all}, we present the prompt templates and object tuples used to construct the prompt sets for each condition in our preliminary analysis~(Section 3).

\begin{center}
  \centering
  \scriptsize
  \resizebox{0.98\textwidth}{!}{%
    \begin{tabular}{@{} l l l p{9cm} @{}}
      \toprule
      \textbf{Aspect} & \textbf{Condition} & \textbf{Prompt Template} & \textbf{Object Tuples} \\
      \midrule
      \multirow{2}{*}{Shape}
        & Dissimilar
        & \multirow{2}{*}{\makecell[lt]{``a/an \{$object~A$\} and \\ a/an \{$object~B$\}’’}}
        & (coin, brick), (donut, chocolate bar), (plate, ice cube), (basketball, television), (balloon, window), (coin, chocolate bar), (donut, soap bar), (basketball, picture frame), (balloon, monitor), (orange, sugar cube), (shoebox, clock), (soap bar, ring), (sugar cube, frisbee), (monitor, orange), (picture frame, bubble), (brick, ring), (ice cube, frisbee), (television, bubble), (picture frame, orange), (brick, bubble) \\
      \cmidrule(lr){2-4}
        & Similar
        & \multirow{2}{*}{\makecell[lt]{``a/an \{$object~A$\} and \\ a/an \{$object~B$\}’’}}
        & (coin, clock), (donut, ring), (plate, frisbee), (basketball, orange), (balloon, bubble), (coin, ring), (donut, frisbee), (basketball, bubble), (balloon, orange), (orange, bubble), (shoebox, brick), (soap bar, chocolate bar), (sugar cube, ice cube), (monitor, television), (picture frame, window), (brick, chocolate bar), (ice cube, soap bar), (television, picture frame), (picture frame, monitor), (brick, sugar cube) \\
      \midrule
      \multirow{2}{*}{Texture}
        & Dissimilar
        & \multirow{2}{*}{\makecell[lt]{``a/an \{$object~A$\} and \\ a/an \{$object~B$\}’’}}
        & (fur coat, coin), (peach, saucepan), (sweater, wrench), (cardigan, belt buckle), (rabbit, kettle), (fur coat, key), (mitten, knife), (sweater, saucepan), (puppy, kettle), (cardigan, wrench), (screw, moss patch), (key, mink coat), (bolt, mitten), (ring, towel), (medal, puppy), (screw, peach), (belt buckle, mink coat), (wrench, moss patch), (ring, peach), (saucepan, sweater) \\
      \cmidrule(lr){2-4}
        & Similar
        & \multirow{2}{*}{\makecell[lt]{``a/an \{$object~A$\} and \\ a/an \{$object~B$\}’’}}
        & (fur coat, moss patch), (peach, mink coat), (sweater, mitten), (cardigan, towel), (rabbit, puppy), (fur coat, peach), (mitten, mink coat), (sweater, towel), (puppy, moss patch), (cardigan, sweater), (screw, coin), (key, saucepan), (bolt, wrench), (ring, belt buckle), (medal, kettle), (screw, key), (belt buckle, saucepan), (wrench, screw), (ring, kettle), (saucepan, wrench) \\
      \midrule
      \multirow{2}{*}{Background Bias}
        & Similar
        & \multirow{2}{*}{\makecell[lt]{``a/an \{$object~A$\} and \\ a/an \{$object~B$\}’’}}
        & (goat, sheep), (monkey, chicken), (lion, deer), (pig, chicken), (cow, monkey), (chicken, lion), (goat, rabbit), (rabbit, gorilla), (chicken, sheep), (monkey, deer), (seahorse, octopus), (shrimp, whale), (dolphin, starfish), (jellyfish, shark), (shark, clownfish), (dolphin, seahorse), (starfish, whale), (seahorse, clownfish), (shrimp, shark), (tuna, shrimp) \\
      \cmidrule(lr){2-4}
        & Dissimilar
        & \multirow{2}{*}{\makecell[lt]{``a/an \{$object~A$\} and \\ a/an \{$object~B$\}’’}}
        & (goat, octopus), (monkey, whale), (lion, starfish), (pig, shark), (cow, octopus), (chicken, seahorse), (goat, whale), (rabbit, clownfish), (chicken, shark), (monkey, shrimp), (seahorse, sheep), (shrimp, chicken), (dolphin, deer), (jellyfish, chicken), (shark, monkey), (dolphin, lion), (starfish, rabbit), (seahorse, gorilla), (shrimp, sheep), (tuna, deer) \\
      \midrule
      \multirow{4}{*}{Object Count}
        & \multirow{2}{*}{Few}
        & \makecell[lt]{``a/an \{$object~A$\}’’}
        & (dog), (cat), (lion), (horse), (penguin), (rabbit), (gorilla), (bear), (cow), (monkey) \\
      \cmidrule(lr){3-4}
        & 
        & \makecell[lt]{``a/an \{$object~A$\} and \\ a/an \{$object~B$\}’’}
        & (dog, cat), (cat, lion), (horse, penguin), (rabbit, gorilla), (penguin, horse), (bear, cow), (horse, lion), (rabbit, cat), (cow, penguin), (dog, lion) \\
      \cmidrule(lr){2-4}
        & \multirow{2}{*}{Many}
        & \makecell[lt]{``a/an \{$object~A$\}, \\ a/an \{$object~B$\}, and \\ a/an \{$object~C$\}’’}
        & (dog, cat, horse), (cat, lion, penguin), (horse, penguin, cat), (rabbit, gorilla, cow), (penguin, horse, gorilla), (bear, cow, horse), (horse, lion, cat), (rabbit, cat, dog), (cow, penguin, dog), (dog, lion, cat) \\
      \cmidrule(lr){3-4}
        & 
        & \makecell[lt]{``a/an \{$object~A$\}, \\ a/an \{$object~B$\}, \\ a/an \{$object~C$\}, and \\ a/an \{$object~D$\}’’}
        & (dog, cat, horse, gorilla), (cat, lion, penguin, bear), (horse, penguin, cat, lion), (rabbit, gorilla, cow, cat), (penguin, horse, gorilla, rabbit), (bear, cow, horse, penguin), (horse, lion, cat, penguin), (rabbit, cat, dog, cow), (cow, penguin, dog, gorilla), (dog, lion, cat, rabbit) \\
      \bottomrule
    \end{tabular}%
  }
  \captionof{table}{Prompt templates and object tuples used to construct the prompt sets for each condition in our preliminary analysis~(Section 3).}
  \label{tab:appendix_condition_group_full_list_all}
\end{center}

]

\clearpage


\section{Experimental Details}
\label{sec:appendix_experimental_details}

\subsection{Implementation details}
\label{sec:appendix_implementation_details}
All experiments are conducted on a single NVIDIA RTX 3090 GPU. Our method builds on SDXL~\cite{PodellEtAl2023} and is also implemented with the recently released Stable Diffusion 3.5~(SD3.5)~\cite{stabilityai2024stable35}. To construct the adaptive strengths defined in Eq.~\ref{eqn:adaptive strength}, we use 42 representative words covering a diverse range of object shapes and textures and 36 representative background phrases, which are selected through interaction with GPT-o4-mini-high~\cite{openai_gpt-o4-mini-high_2025}. The full lists of these words and phrases are shown in Table~\ref{tab:appendix_adaptive_strength}. Table~\ref{tab:appendix_offsets} presents the offsets used in the shifted tempered sigmoid function in Eq.~\ref{eqn:adaptive strength}, provided for each base model and embedding type. All offsets are precomputed over all object pairs from 80 MS-COCO~\cite{lin2014microsoft} classes.

\begin{table}[ht]
  \centering
  \footnotesize
  \setlength{\tabcolsep}{3pt}
  \resizebox{\columnwidth}{!}{%
    \begin{tabular}{@{} l p{0.95\columnwidth} @{}}
      \toprule
      Category   & Selected Words or Phrases  \\
      \midrule
      Shape or Texture  &
        round, oval, square, rectangular, triangular, spherical, cylindrical, conical, flat, elongated, pointed, curved, ring‑shaped, disc‑shaped, irregular, spiral, star‑shaped, jagged, boxy, stocky, smooth, rough, bumpy, grainy, granular, fuzzy, hairy, woolly, leathery, scaly, feathery, slimy, wrinkled, shaggy, shiny, matte, porous, spongy, striped, spotted, patterned, rubbery \\
      \cmidrule(lr){1-2}
      Background &
        in a forest, in a desert, on a mountain, on a beach, in a grassland, in the Arctic, in the savanna, on the ocean surface, underwater, in a river, in a lake, in a coral reef, in a cave, in a city street, in a suburban neighborhood, in a farm field, at a construction site, at a stadium, in a parking lot, on a road, on a railway track, on a boat deck, on an airport runway, in the sky, in space, in a living room, in a kitchen, in a bedroom, in a bathroom, in an office, in a classroom, in a laboratory, in a factory, in a warehouse, in a shopping mall, in a grocery store \\
      \bottomrule
    \end{tabular}%
  }
  \caption{Lists of 42 representative words describing object shapes and textures and 36 representative background phrases, used to construct attribute and background prompts for computing adaptive strengths defined in Eq.~\ref{eqn:adaptive strength}.}
  \label{tab:appendix_adaptive_strength}
\end{table}
\begin{table}[ht]
  \centering
  \scriptsize
    \begin{tabular}{llccc}
      \toprule
      Base Model & Offset & \multicolumn{3}{c}{Embedding Type~($\tau$)} \\
      \cmidrule(lr){3-5}
      &         & $\mathrm{obj}$  & $\mathrm{EOT}$    & $\mathrm{pool}$ \\
      \midrule
      \multirow{2}{*}{SDXL}
        & $x^{\mathrm{attr}}_{\tau,0}$ & 0.5550 & 0.5474 & 0.5366 \\
        & $x^{\mathrm{bg}}_{\tau,0}$   & 0.1592 & 0.3862 & 0.5835 \\
      \cmidrule(l){1-5}
      \multirow{2}{*}{SD3.5}
        & $x^{\mathrm{attr}}_{\tau,0}$ & 0.5536 & 0.5473 & 0.6168 \\
        & $x^{\mathrm{bg}}_{\tau,0}$   & 0.1705 & 0.3877 & 0.4325 \\
      \bottomrule
    \end{tabular}%
  \caption{Offsets \(x^{\mathrm{attr}}_{\tau,0}\) and \(x^{\mathrm{bg}}_{\tau,0}\) in Eq.~\ref{eqn:adaptive strength} for each base model~(SDXL, SD3.5) and each embedding type $\tau \in \{ \mathrm{obj, EOT, pool}\}$.}
  \label{tab:appendix_offsets}
\end{table}

\subsection{Benchmark details}
\label{sec:appendix_benchmark_details}
We construct four benchmarks corresponding to the problematic scenarios reported in Figure 1, which are used in our main experiments: Similar Shapes, Similar Textures, and Dissimilar Background Biases, and Many Objects. Each of the first three benchmarks consists of 50 two-object prompts, following the template of ``a/an $\{object~A\}$ and a/an $\{object~B\}$.'' The object pairs for these benchmarks are selected through interactions with GPT-o4-mini-high~\cite{openai_gpt-o4-mini-high_2025}. For example, to obtain 50 object pairs for the Similar Shapes benchmark, we ask the model: ``Suggest 50 pairs of objects that share a similar overall shape or silhouette, even if they are unrelated in nature.'' For the Many Objects benchmark, we extend the animal list used in Attend-and-Excite to include 17 animals, from which we randomly sample to create 25 three-object prompts and 25 four-object prompts. The extended animal list is defined as [cat, dog, bird, bear, lion, horse, elephant, monkey, frog, turtle, rabbit, fish, panda, cow, gorilla, penguin, chicken]. Table~\ref{tab:appendix_full_benchmark} shows the prompt templates and object tuples used to construct 50 prompts for each benchmark.

\clearpage

\begin{table*}[p]
  \centering
  \setlength{\tabcolsep}{4pt}
  \resizebox{\textwidth}{!}{%
    \begin{tabular}{@{} l p{6cm} p{11cm} @{}}
      \toprule
      \textbf{Benchmark} & \textbf{Prompt Template} & \textbf{Object Tuples} \\
      \midrule
      Similar Shapes
        & ``a/an \{$object~A$\} and a/an \{$object~B$\}''
        & (basketball, orange), (balloon, ball), (coin, button), (mushroom, umbrella), (soap, eraser), (bicycle, motorcycle), (plate, frisbee), (crayon, candle), (soda can, battery), (golf club, hockey stick), (tent, pyramid), (traffic cone, party hat), (carrot, ice cream cone), (snake, rope), (leaf, feather), (horseshoe, magnet), (soccer ball, globe), (light bulb, onion), (jellyfish, parachute), (butterfly, bow tie), (tennis ball, lime), (hedgehog, hairbrush), (coin, clock), (remote control, chocolate bar), (comb, rake), (jellybean, kidney bean), (matchstick, pencil), (marker, lipstick), (donut, ring), (pebble, almond), (jellyfish, octopus), (penguin, bowling pin), (donut, compact disc), (cabbage, balloon), (pencil, straw), (broom, spear), (screwdriver, paintbrush), (onion, egg), (fork, trident), (shovel, oar), (battery, bullet), (coin, medal), (hourglass, dumbbell), (car tire, lifebuoy), (paperclip, pretzel), (rocket, carrot), (donut, lifebuoy), (clock, wheel), (balloon, bubble), (book, brick) \\
      \cmidrule(lr){1-3}
      Similar Textures
        & ``a/an \{$object~A$\} and a/an \{$object~B$\}''
        & (zebra, referee shirt), (leopard, giraffe), (diamond, ice cube), (belt, wallet), (waffle, honeycomb), (coral, sponge), (popcorn, cauliflower), (sandcastle, sugar cube), (loaf of bread, cork), (cactus, pineapple), (marble, ice cube), (sock, mitten), (fur coat, moss patch), (soap, candle), (key, spoon), (strawberry, golf ball), (jellyfish, plastic bag), (kiwi, coconut), (soccer ball, turtle shell), (cactus, sea urchin), (crab, shrimp), (bottle, jar), (fork, spoon), (sheep, cotton ball), (kiwi bird, coconut), (tarantula, carpet), (crocodile, leather handbag), (tiger, clownfish), (panda, soccer ball), (penguin, tuxedo), (giraffe, cheetah), (mirror, chrome ball), (coin, spoon), (carpet, rabbit), (mink coat, peach), (leopard, dalmatian), (teddy bear, peach), (fish, sequined dress), (sponge, pumice stone), (leather shoe, basketball), (cherry, grape), (olive, grape), (wine glass, light bulb), (ladybug, dalmatian), (cheetah, ladybug), (pangolin, artichoke), (lobster, shrimp), (sponge, swiss cheese), (brick, sandpaper), (cat, rabbit) \\
      \cmidrule(lr){1-3}
      \makecell[lt]{Dissimilar \\ Background Biases}
        & ``a/an \{$object~A$\} and a/an \{$object~B$\}''
        & (beach ball, snowball), (coconut, ice cube), (sandcastle, igloo), (kayak, snowboard), (camel, seahorse), (cow, whale), (goat, octopus), (chicken, fish), (chameleon, seahorse), (horse, seal), (chameleon, penguin), (rabbit, seal), (pineapple, coral), (apple, clam), (coconut, starfish), (lettuce, seaweed), (cauliflower, seahorse), (potato, oyster), (lion, whale), (camel, polar bear), (elephant, shark), (cactus, penguin), (penguin, camel), (fish, bicycle), (crocodile, eagle), (deer, shark), (rabbit, crab), (eagle, seahorse), (igloo, desert tent), (snowman, sandcastle), (ice skate, surfboard), (crocodile, camel), (koala, dolphin), (bear, crab), (cat, jellyfish), (giraffe, seahorse), (sofa, tent), (tractor, sailboat), (boat, car), (cactus, seahorse), (cactus, coral), (strawberry, fish), (parrot, penguin), (tank, canoe), (polar bear, kangaroo), (deer, sea turtle), (monkey, octopus), (carrot, seaweed), (carrot, coral), (dog, dolphin) \\
      \cmidrule(lr){1-3}
      \multirow{2}{*}{Many Objects}
        & ``a/an \{$object~A$\}, a/an \{$object~B$\}, and a/an \{$object~C$\}''
        & (penguin, cat, elephant), (panda, frog, horse), (bear, frog, fish), (lion, fish, chicken), (cat, horse, gorilla), (chicken, bear, rabbit), (gorilla, horse, bird), (cat, fish, turtle), (frog, monkey, fish), (panda, gorilla, bird), (dog, rabbit, lion), (cow, panda, turtle), (chicken, bear, monkey), (rabbit, fish, monkey), (bear, fish, bird), (cow, fish, cat), (fish, bear, cat), (chicken, bird, bear), (chicken, lion, cat), (chicken, bear, penguin), (dog, horse, bird), (monkey, turtle, chicken), (cow, gorilla, bird), (lion, turtle, monkey), (horse, frog, fish) \\
      \cmidrule(lr){2-3}
        & ``a/an \{$object~A$\}, a/an \{$object~B$\}, a/an \{$object~C$\}, and a/an \{$object~D$\}''
        & (bear, horse, turtle, frog), (chicken, cat, bear, lion), (bird, cow, horse, turtle), (turtle, fish, horse, dog), (penguin, cat, cow, gorilla), (fish, horse, gorilla, penguin), (cow, chicken, monkey, turtle), (turtle, horse, cow, gorilla), (bird, cat, fish, cow), (rabbit, turtle, cat, penguin), (bird, cat, dog, elephant), (fish, turtle, chicken, frog), (lion, frog, rabbit, fish), (frog, chicken, rabbit, fish), (chicken, frog, monkey, dog), (chicken, elephant, bird, frog), (cat, panda, horse, bear), (turtle, cat, frog, fish), (cat, cow, horse, monkey), (cow, chicken, turtle, rabbit), (fish, bird, bear, turtle), (cat, lion, rabbit, dog), (bear, cat, penguin, chicken), (panda, fish, chicken, monkey), (panda, lion, frog, chicken) \\
      \bottomrule
    \end{tabular}%
  }
  \caption{Prompt templates and object tuples for the four benchmarks evaluated in our main experiments.}
  \label{tab:appendix_full_benchmark}
\end{table*}

\clearpage

\subsection{Evaluation metrics}
\label{sec:appendix_evaluation_metrics}
\begin{figure}[ht]
  \centering
  \includegraphics[width=\columnwidth]{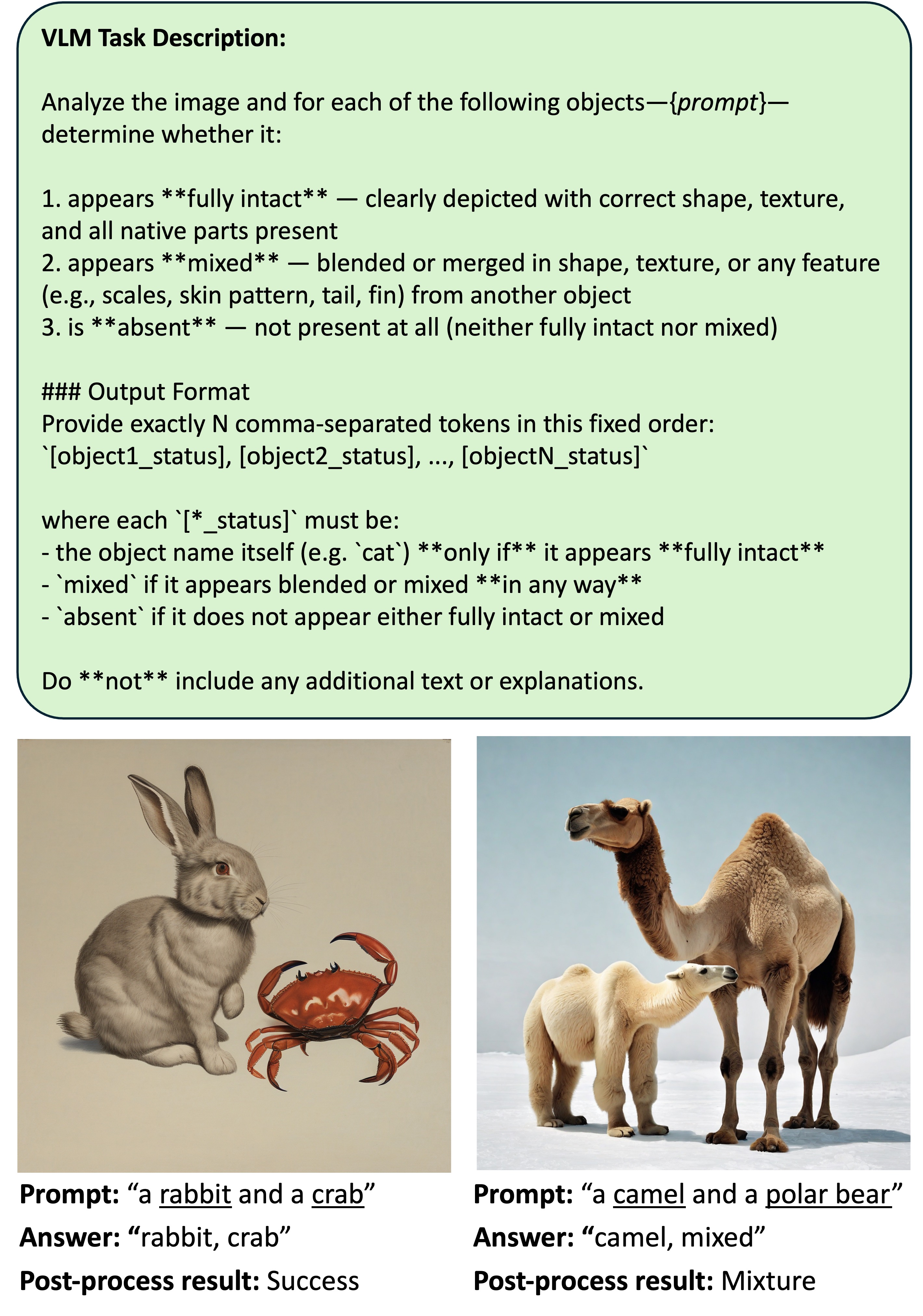}
  \caption{Examples illustrating our VLM-based evaluation.} 
  \label{fig:vlm_example}
\end{figure}

We adjust the VLM-based metric from EnMMDiT~\cite{wei2024enhancing} to measure both the success rate~(SR) and mixture rate~(MR) for each generated image. We provided GPT-4o-mini with the generated image and the text prompt used for image generation, asking it to classify each object in the prompt as (1) fully intact, (2) mixed, or (3) absent. SR is calculated as the ratio of images where all objects are classified as fully intact, and MR is calculated as the ratio of images where one or more objects are classified as mixed. Figure~\ref{fig:vlm_example} shows examples illustrating our VLM-based evaluation.


\subsection{Human preference study}
\label{sec:appendix_human_evaluation}
\begin{figure}[ht]
  \centering
  \includegraphics[width=\columnwidth]{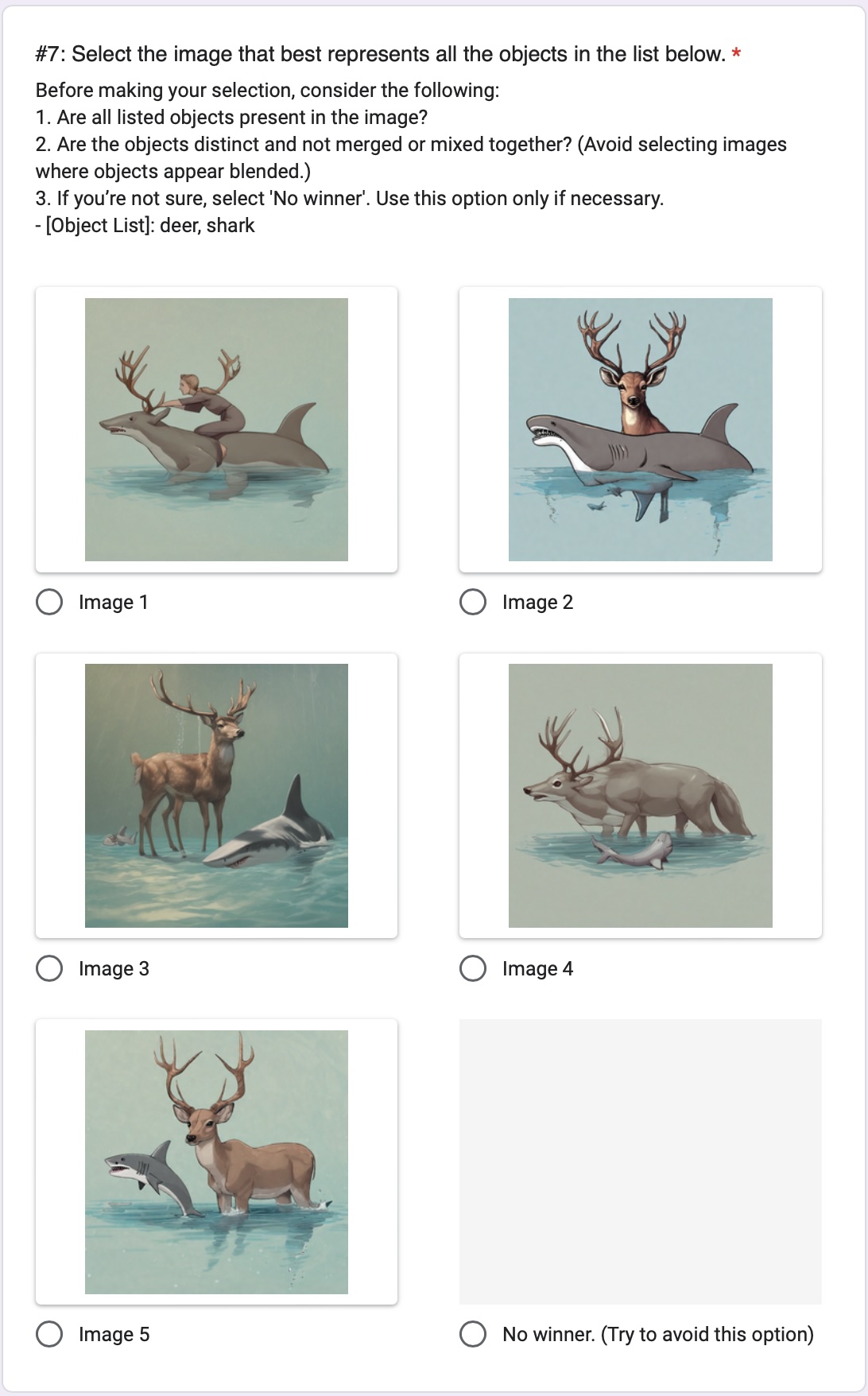}
  \caption{A screenshot of a question used in our human preference studies.}
  \label{fig:human_evaluation}
\end{figure}

For the human preferences studies in Section~\ref{subsec:Experimental Results}, we used Amazon Mechanical Turk (MTurk) to collect responses, requiring participants to have over 500 HIT approvals, an approval rate above 98\%, and US residency. In each survey, participants were shown five images generated from the same text prompt but using different SDXL-based methods, and asked to select the one that best represents all the objects specified in the prompt. Participants were instructed to avoid images with object neglect or object mixing. Each participant answered 20 questions, along with one additional dummy question designed to assess attentiveness. The correct answer to the dummy question was provided at the beginning of each survey, and only participants who answered it correctly were included in the valid set. As a result, we collected 580, 640, 640, 560 responses from 29, 32, 32, 28 valid participants for each survey, respectively. Figure~\ref{fig:human_evaluation} shows a screenshot of a question used in our human preference studies.

\clearpage


\section{Additional Experiments}
\label{sec:appendix_additional_experiments}

\subsection{Combining DOS with Attend-and-Excite}
\label{sec:appendix_incorporate_with_other_methods}
\begin{figure}[ht]
  \centering
  \includegraphics[width=\columnwidth]{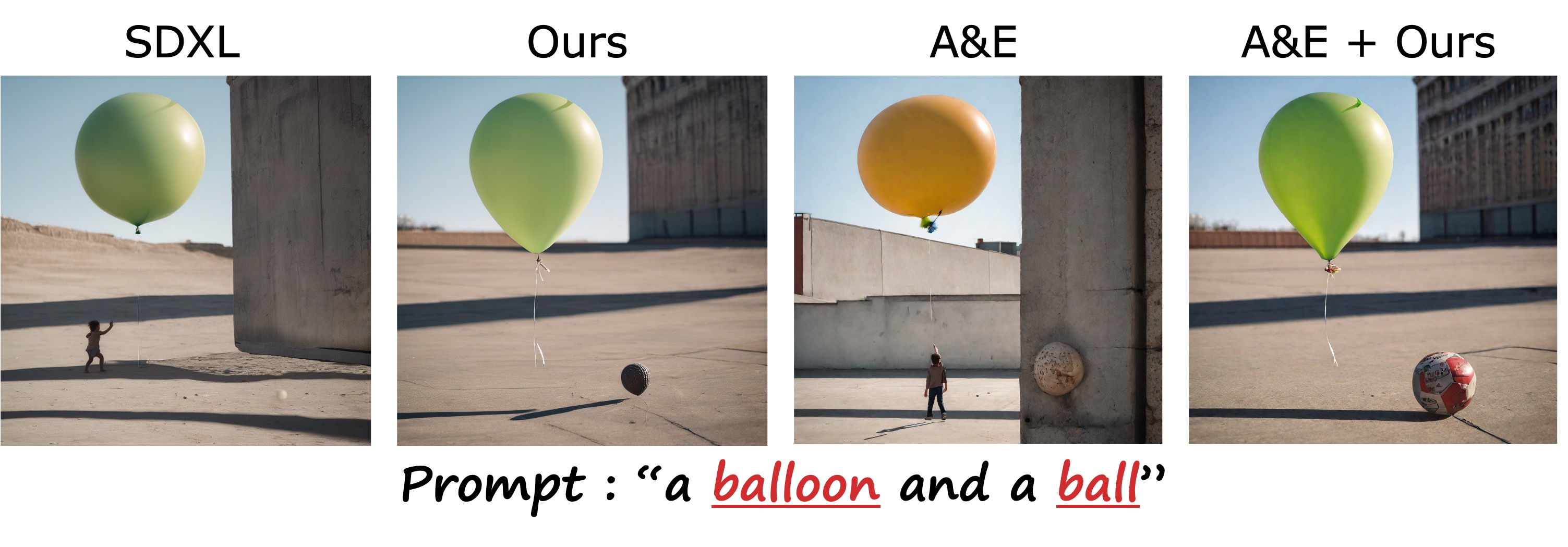}
  \caption{Our method, DOS, can be used in combination with other latent modification method such as A\&E~\citep{chefer2023attend}. All methods are based on SDXL.}
  \label{fig:incorporate_with_other_methods}
\end{figure}

Since DOS only modifies CLIP text embeddings, it can be used in combination with other latent modification methods such as Attend-and-Excite~(A\&E)~\citep{chefer2023attend}. This combined method addresses some issues that cannot be solved by using either DOS or A\&E alone, as shown in Figure~\ref{fig:incorporate_with_other_methods}. We further validate the effectiveness of this combination through a quantitative comparison, presented in Table~\ref{tab:appendix_combine_with_ae}. 

\begin{table}[ht]
  \centering
  \small
  \setlength{\tabcolsep}{2pt} 
  \resizebox{\columnwidth}{!}{%
    \begin{tabular}{@{}l cc cc cc cc@{}}
      \toprule
      \textbf{Method}
        & \multicolumn{2}{c}{\makecell{Similar\\Shapes}}
        & \multicolumn{2}{c}{\makecell{Similar\\Textures}}
        & \multicolumn{2}{c}{\makecell{Dissimilar\\Background Biases}}
        & \multicolumn{2}{c}{\makecell{Many\\Objects}} \\
      \cmidrule(lr){2-3} \cmidrule(lr){4-5} \cmidrule(lr){6-7} \cmidrule(lr){8-9}
        & SR$\uparrow$ & MR$\downarrow$
        & SR$\uparrow$ & MR$\downarrow$
        & SR$\uparrow$ & MR$\downarrow$
        & SR$\uparrow$ & MR$\downarrow$ \\
      \midrule
      SDXL
        & 48.0\% & 6.5\%
        & 58.0\% & 7.5\%
        & 46.0\% & 22.5\%
        & 23.0\% & 27.5\% \\
      Ours
        & 64.0\% & 3.5\%
        & 71.5\% & 3.5\%
        & 68.5\% & 17.0\%
        & 48.0\% & 15.5\% \\
      A\&E
        & 60.5\% & 6.0\%
        & 67.5\% & 5.0\%
        & 53.5\% & 25.0\%
        & 28.5\% & 28.5\% \\
      A\&E+Ours
        & \textbf{66.0\%} & \textbf{3.0\%}
        & \textbf{73.5\%} & \textbf{3.0\%}
        & \textbf{69.0\%} & \textbf{14.5\%}
        & \textbf{52.5\%} & \textbf{15.0\%} \\
      \bottomrule
    \end{tabular}%
  }
  \caption{The combination of DOS and A\&E (A\&E+Ours) demonstrates a synergistic effect, achieving the best performance across all benchmarks.}
  \label{tab:appendix_combine_with_ae}
\end{table}

\subsection{Qualitative ablation on text embedding types}
\label{sec:appendix_qualitative_ablation_on_text_embedding_types}
\begin{figure}[ht]
  \centering
  \includegraphics[width=\columnwidth]{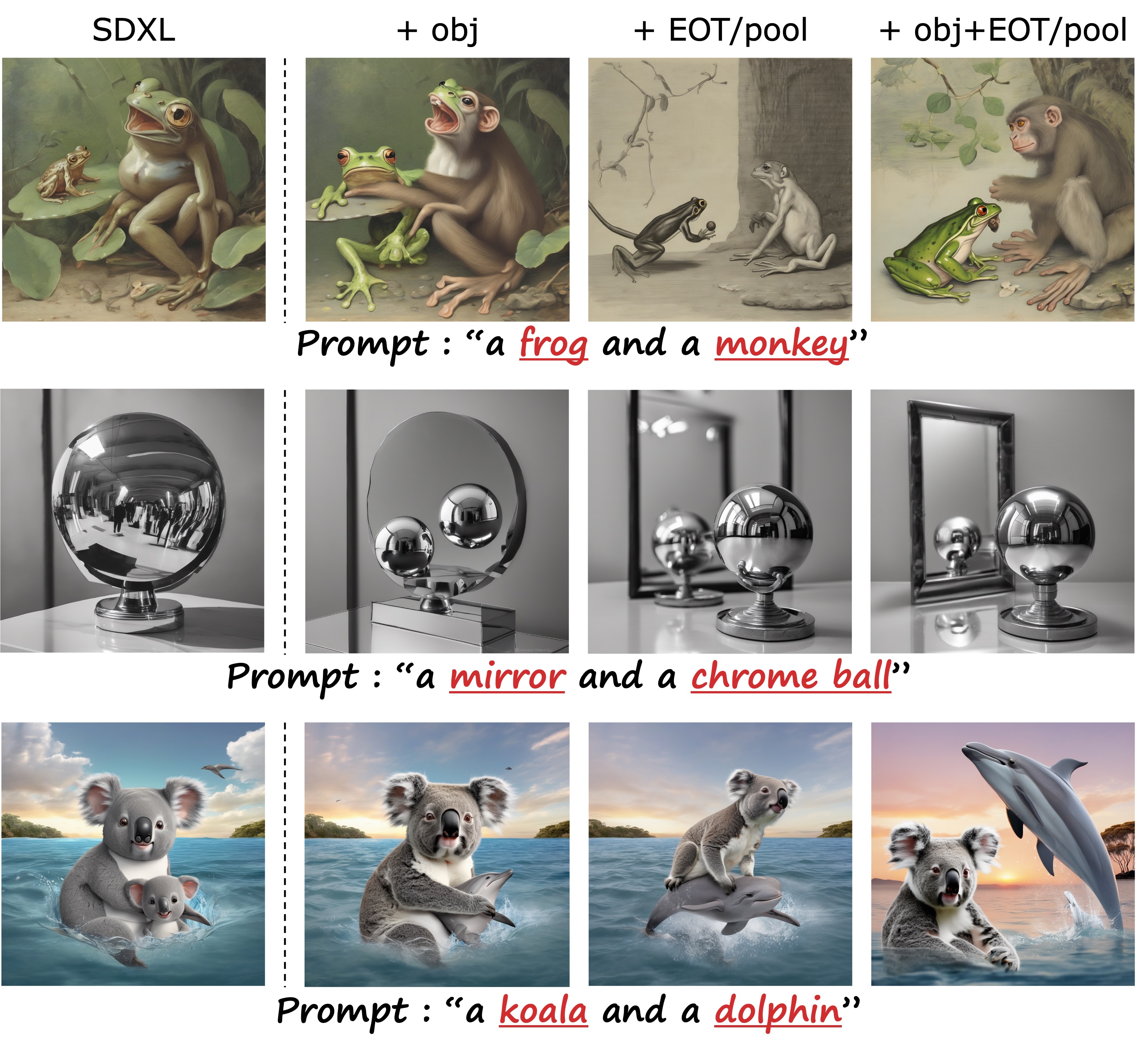}
  \caption{
  Qualitative results from the ablation study on the types of text embeddings to which DOS is applied. Applying DOS to all types of text embeddings produces the best outcomes, clearly representing both objects in the prompt without object mixing.
    }
  \label{fig:ablation_qualitative}
\end{figure}

Figure~\ref{fig:ablation_qualitative} presents qualitative examples from the ablation study on the types of text embeddings to which DOS is applied. Applying DOS only to semantic token embeddings~(+obj) or only to EOT/pooled embeddings~(+EOT/pool) yields suboptimal results, often exhibiting object neglect or mixing. In contrast, applying DOS to all types of text embeddings produces the best outcomes, clearly representing both objects in the prompt without object mixing.





\begin{figure*}[ht]
  \centering
  \includegraphics[width=0.95\textwidth]{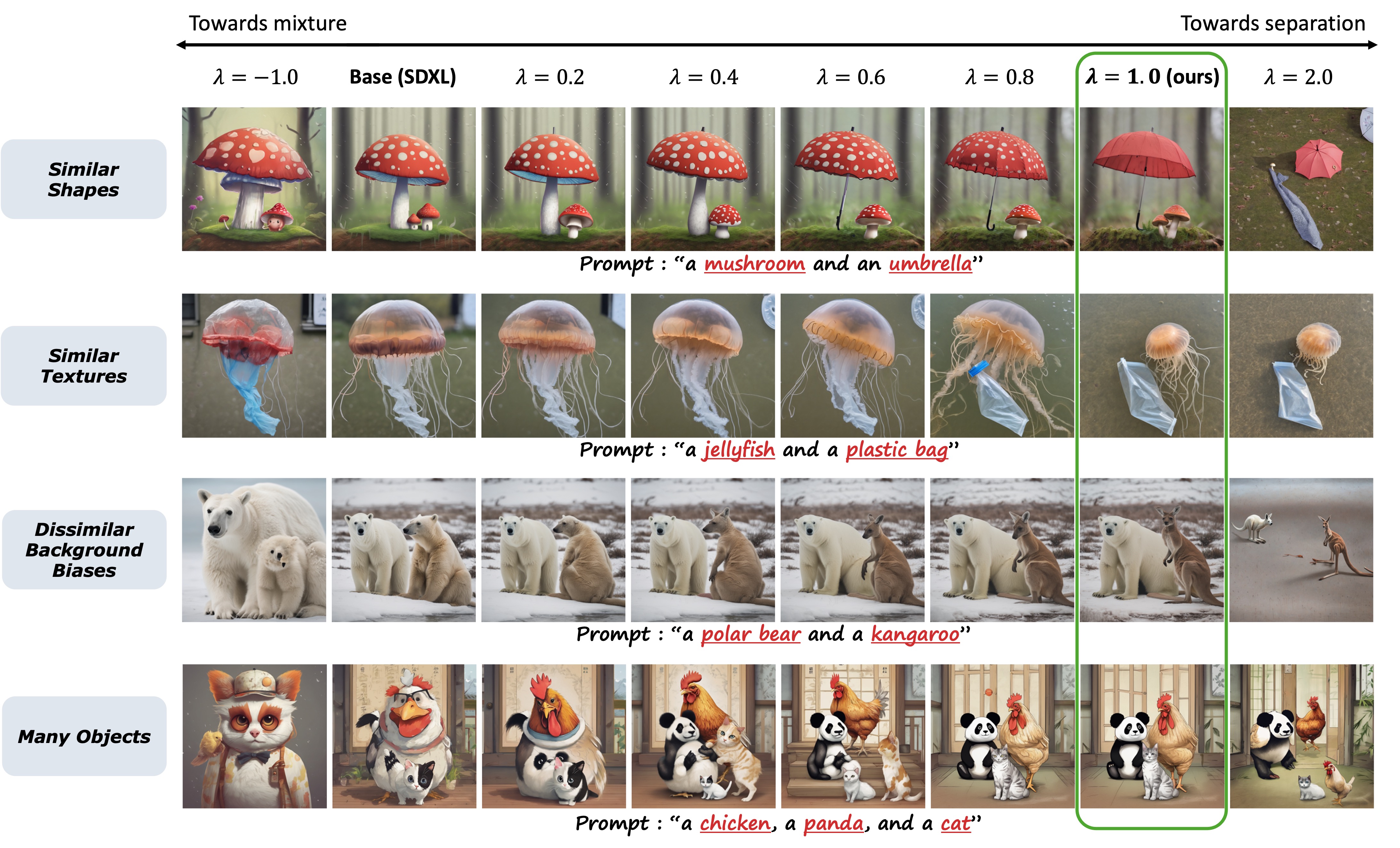}
  \captionof{figure}{Directional information encoded in DOS vectors. Effect of scaling the directional scaler $\lambda$ defined in Equations~\ref{eqn:appendix_scale_obj}-\ref{eqn:appendix_scale_eot_pool}. As $\lambda$ increases positively, objects become more separated, while the negative value of $\lambda$ intensifies object mixing. Excessively large positive values (i.e., $\lambda=2$) can cause unexpected results due to strong deviation from the original text embeddings. The baseline SDXL corresponds to using $\lambda=0$, while our default method corresponds to using $\lambda=1$.}
  \label{fig:scaling_DOS}
\end{figure*}

\subsection{Extended Many Objects benchmark}
\label{sec:extend_object_categories}

To address concerns regarding the benchmark's scope and to validate the external validity of our method, we construct an extended Many Objects benchmark with greater categorical diversity. Moving beyond the original animal-only set, we incorporate objects from electronics, vehicles, and fruits. The expanded object list now consists of 16 distinct objects: [cat, dog, frog, fish, smartphone, laptop, television, camera, car, bicycle, motorcycle, train, apple, orange, avocado, strawberry]. From this set, we randomly generate 25 three-object and 25 four-object prompts. As shown in Table~\ref{tab:appendix_many_objects_v2}, the performance trends are consistent with those on the original benchmark presented in Table~\ref{tab:performance_comparison}. This result indicates that our method's effectiveness is not limited to a single category but generalizes across diverse object categories.

\begin{table}[ht]
  \centering
  \scriptsize
  \begin{tabular}{llcc}
    \toprule
    \ Base Model & Method  & SR$\uparrow$ & MR$\downarrow$ \\
    \midrule
    \multirow{5}{*}{SDXL}  & Base             & 31.50\%          & 19.50\%          \\
                           & TEBOpt           & 33.50\%          & 21.50\%          \\
                           & A\&E             & 42.50\%          & 18.50\%          \\
                           & CONFORM          & 43.50\%          & 21.00\%          \\
                           & Ours             & \textbf{45.00\%} & \textbf{14.50\%} \\
    \midrule 
    \multirow{2}{*}{SD3.5} & Base             & 78.00\%          & 13.00\%          \\
                           & Ours             & \textbf{85.50\%} & \textbf{6.00\%}  \\
    \bottomrule
  \end{tabular}
  \caption{Performance comparison on the extended Many Objects benchmark. SR denotes Success Rate~(↑ higher is better), and MR denotes Mixture Rate~(↓ lower is better). The best results are highlighted in bold.}
  \label{tab:appendix_many_objects_v2}
\end{table}

\subsection{Sensitivity to predefined words and phrases}
\label{sec:appendix_sensitivity_ablation}

Our adaptive strength utilizes predefined lists of 42 shape/texture words and 36 background phrases. To ensure our approach is not over-tuned to this specific set, we performed a sensitivity analysis. We created a reduced set by randomly sampling half of these lists (21 words and 18 phrases). We then evaluated the performance of DOS on this reduced set, repeating the process four times and averaging the outcomes. The results, shown in Table~\ref{tab:appendix_sensitivity_adaptive_strength}, reveal that while there is a slight drop in performance, the method's effectiveness is largely maintained. This demonstrates the robustness of our adaptive mechanism, alleviating concerns about its dependency on a fixed, hand-crafted set of priors.

\begin{table}[t]
  \centering
  \small
  \setlength{\tabcolsep}{2pt} 
  \resizebox{\columnwidth}{!}{%
    \begin{tabular}{@{}l cc cc cc cc@{}}
      \toprule
      \textbf{Method}
        & \multicolumn{2}{c}{\makecell{Similar\\Shapes}}
        & \multicolumn{2}{c}{\makecell{Similar\\Textures}}
        & \multicolumn{2}{c}{\makecell{Dissimilar\\Background Biases}}
        & \multicolumn{2}{c}{\makecell{Many\\Objects}} \\
      \cmidrule(lr){2-3} \cmidrule(lr){4-5} \cmidrule(lr){6-7} \cmidrule(lr){8-9}
        & SR$\uparrow$ & MR$\downarrow$
        & SR$\uparrow$ & MR$\downarrow$
        & SR$\uparrow$ & MR$\downarrow$
        & SR$\uparrow$ & MR$\downarrow$ \\
      \midrule
      SDXL
        & 48.0\% & 6.5\%
        & 58.0\% & 7.5\%
        & 46.0\% & 22.5\%
        & 23.0\% & 27.5\% \\
      Ours (50\% words/phrases)
        & 62.4\% & \textbf{2.4\%}
        & 69.4\% & 4.8\%
        & 67.4\% & 18.8\%
        & 47.8\% & \textbf{13.9\%} \\
      Ours (100\% words/phrases)
        & \textbf{64.0\%} & 3.5\%
        & \textbf{71.5\%} & \textbf{3.5\%}
        & \textbf{68.5\%} & \textbf{17.0\%}
        & \textbf{48.0\%} & 15.5\% \\
      \bottomrule
    \end{tabular}%
  }
  \caption{Ablation study on the predefined words and phrases. Performance is compared between using the full list of predefined words and phrases (100\%) and a randomly sampled half (50\%).}
  \label{tab:appendix_sensitivity_adaptive_strength}
\end{table}

\subsection{DOS vectors encode directional information}
\label{sec:appendix_linearity_of_semantic_directions}

DOS vectors are weighted sum of separation vectors, where each separation vector reflects the directional information for separating the corresponding object pair. Therefore, DOS vectors may also carry directional information, which can be observed in the generated images. To investigate this, we vary the directional scale of each DOS vector before adding them to the original CLIP text embeddings as follows:
\begin{align}
  {\bm{c}_{\mathrm{obj}}^n}'      &= \bm{c}_{\mathrm{obj}}^{n}
    + \lambda\,\mathbf{v}^{\mathrm{DOS},n}_{\mathrm{obj}},
    \label{eqn:appendix_scale_obj}
  \\
  \bm{c}^{\prime}_{\mathrm{EOT/pool}} &= \bm{c}_{\mathrm{EOT/pool}}
    + \lambda\sum_{i=1}^{N}\mathbf{v}^{\mathrm{DOS},i}_{\mathrm{EOT/pool}},
    \label{eqn:appendix_scale_eot_pool}
\end{align}
where $n \in {1, \dots, N}$ denotes the index of an object, and $\lambda$ is a directional scaler (which can also take negative values) (Please compare these with Equations~\ref{eqn:DOS_update_semantic token embeddings}–\ref{eqn:DOS_update_semantic EOT/pooled embeddings}). Figure~\ref{fig:scaling_DOS} shows generated images with $\lambda$ varying from $-1.0$ to $2.0$ for four prompts. As $\lambda$ increases in the positive direction, the objects in the prompts gradually become more separated. In contrast, when $\lambda$ is negative, the objects tend to mix more. Lastly, increasing $\lambda$ too much in the positive direction (e.g., $\lambda=2$) occasionally leads to unexpected results, possibly due to a strong deviation from the original text embeddings.

\begin{table*}[t]
  \centering
  \scriptsize
  \setlength{\tabcolsep}{3pt} 
  \resizebox{\textwidth}{!}{%
    \begin{tabular}{@{}ll ccc ccc ccc ccc@{}}
      \toprule
      \multirow{2}{*}{Base Model} & \multirow{2}{*}{Method}
        & \multicolumn{3}{c}{Similar Shapes}
        & \multicolumn{3}{c}{Similar Textures}
        & \multicolumn{3}{c}{Dissimilar Background Biases}
        & \multicolumn{3}{c}{Many Objects} \\
      \cmidrule(lr){3-5} \cmidrule(lr){6-8} \cmidrule(lr){9-11} \cmidrule(lr){12-14}
      & 
        & \makecell{SR↑ \\ (gemini)} & \makecell{MR↓ \\ (gemini)} & \makecell{SR↑ \\ (g-dino)}
        & \makecell{SR↑ \\ (gemini)} & \makecell{MR↓ \\ (gemini)} & \makecell{SR↑ \\ (g-dino)}
        & \makecell{SR↑ \\ (gemini)} & \makecell{MR↓ \\ (gemini)} & \makecell{SR↑ \\ (g-dino)}
        & \makecell{SR↑ \\ (gemini)} & \makecell{MR↓ \\ (gemini)} & \makecell{SR↑ \\ (g-dino)} \\
      \midrule
      \multirow{5}{*}{SDXL}
        & Base    & 45.50\% & 16.00\% & 24.00\% & 58.00\% & 13.50\% & 39.50\% & 48.00\% & 26.50\% & 43.00\% & 21.50\% & 33.00\% & 17.50\% \\
        & TEBOpt  & 48.50\% & 20.50\% & 24.50\% & 59.00\% & 11.00\% & 37.50\% & 46.00\% & 27.00\% & 40.50\% & 22.00\% & 36.00\% & 22.00\% \\
        & A\&E    & 50.50\% & 14.50\% & 29.50\% & 65.50\% & 9.00\%  & 45.50\% & 62.00\% & 21.00\% & 46.00\% & 33.50\% & 34.50\% & 22.50\% \\
        & CONFORM & 47.50\% & 14.00\% & 24.50\% & 67.00\% & 10.50\% & 44.00\% & 54.00\% & 23.50\% & 47.00\% & 32.50\% & 31.00\% & 27.00\% \\
        & Ours    & \textbf{54.00\%} & \textbf{12.50\%} & \textbf{32.50\%} & \textbf{68.00\%} & \textbf{8.00\%} & \textbf{53.50\%} & \textbf{69.00\%} & \textbf{19.50\%} & \textbf{61.50\%} & \textbf{46.50\%} & \textbf{15.00\%} & \textbf{43.50\%} \\
      \midrule
      \multirow{2}{*}{SD3.5}
        & Base    & 73.00\% & 7.00\% & 38.00\% & 84.50\% & 10.00\% & 64.50\% & 88.50\% & 10.50\% & 72.50\% & 67.00\% & 20.00\% & 59.50\% \\
        & Ours    & \textbf{78.00\%} & \textbf{6.00\%} & \textbf{44.50\%} & \textbf{87.50\%} & \textbf{7.00\%} & \textbf{70.00\%} & \textbf{92.00\%} & \textbf{7.50\%} & \textbf{75.50\%} & \textbf{75.50\%} & \textbf{12.00\%} & \textbf{70.00\%} \\
      \bottomrule
    \end{tabular}%
  }
  \caption{Quantitative comparison using multiple evaluators. We report SR$\uparrow$ (gemini), MR$\downarrow$ (gemini), and SR$\uparrow$ (g-dino) across all four benchmarks. The best results are highlighted in bold.}
  \label{tab:appendix_re-evaluation}
\end{table*}

\subsection{Validation with multiple evaluators}
\label{sec:appendix_validation_metrics}

While the Success Rate (SR) and Mixture Rate (MR) in Table~\ref{tab:performance_comparison} were measured using GPT-4o-mini~\cite{openai_gpt-4o-mini_2024}, relying on a single Vision Language Model (VLM) could introduce evaluation biases. To validate the reliability of our evaluation results, we re-evaluated all methods using two additional evaluators: another VLM, Gemini-2.0-Flash-Lite~\cite{google_gemini-2.0-flash-lite_2025}, and an open-vocabulary object detector, Grounding-DINO~\cite{liu2024grounding}. The MR metric was excluded for the object detector as it is not capable of assessing object mixing. The re-evaluated quantitative comparison, presented in Table~\ref{tab:appendix_re-evaluation}, shows a consistent trend with our main results, suggesting their validity.

\subsection{Performance comparison on SD3.5}
\label{sec:appendix_comparison_on_sd3_5}

Our main comparison in Table~\ref{tab:performance_comparison} was conducted on SDXL as the base model. To verify that this performance advantage extends to more recent architectures, we compared our method against TEBOpt~\cite{chen2024cat} and Self-Cross~\cite{qiu2025self} using SD3.5 as the base model. Table~\ref{tab:appendix_sd3-5-comparison} shows that our approach maintains its superior performance on this newer architecture.

\begin{table}[t]
  \centering
  \small
  \setlength{\tabcolsep}{2pt} 
  \resizebox{\columnwidth}{!}{%
    \begin{tabular}{@{}l cc cc cc cc@{}}
      \toprule
      \textbf{Method}
        & \multicolumn{2}{c}{\makecell{Similar\\Shapes}}
        & \multicolumn{2}{c}{\makecell{Similar\\Textures}}
        & \multicolumn{2}{c}{\makecell{Dissimilar\\Background Biases}}
        & \multicolumn{2}{c}{\makecell{Many\\Objects}} \\
      \cmidrule(lr){2-3} \cmidrule(lr){4-5} \cmidrule(lr){6-7} \cmidrule(lr){8-9}
        & SR$\uparrow$ & MR$\downarrow$
        & SR$\uparrow$ & MR$\downarrow$
        & SR$\uparrow$ & MR$\downarrow$
        & SR$\uparrow$ & MR$\downarrow$ \\
      \midrule
      SD3.5
        & 75.5\% & 4.0\%
        & 79.0\% & 6.0\%
        & 78.0\% & 17.5\%
        & 70.0\% & 16.5\% \\
      TEBOpt
        & 77.5\% & 3.0\%
        & 78.0\% & 8.0\%
        & 81.0\% & 16.5\%
        & 73.0\% & 14.5\% \\
      Self-Cross
        & 77.0\% & \textbf{2.5\%}
        & 81.0\% & 8.0\%
        & 82.5\% & \textbf{13.5\%}
        & 71.5\% & 12.0\% \\
      Ours
        & \textbf{81.0\%} & 3.0\%
        & \textbf{87.5\%} & \textbf{3.0\%}
        & \textbf{85.5\%} & \textbf{13.5\%}
        & \textbf{76.5\%} & \textbf{10.5\%} \\
      \bottomrule
    \end{tabular}%
  }
  \caption{Performance comparison on SD3.5 as the base model.}
  \label{tab:appendix_sd3-5-comparison}
\end{table}

\subsection{Additional qualitative comparisons}
\label{sec:appendix_additional_qualitative_results}

\begin{figure*}[t]
  \centering
  \includegraphics[width=0.7\textwidth]{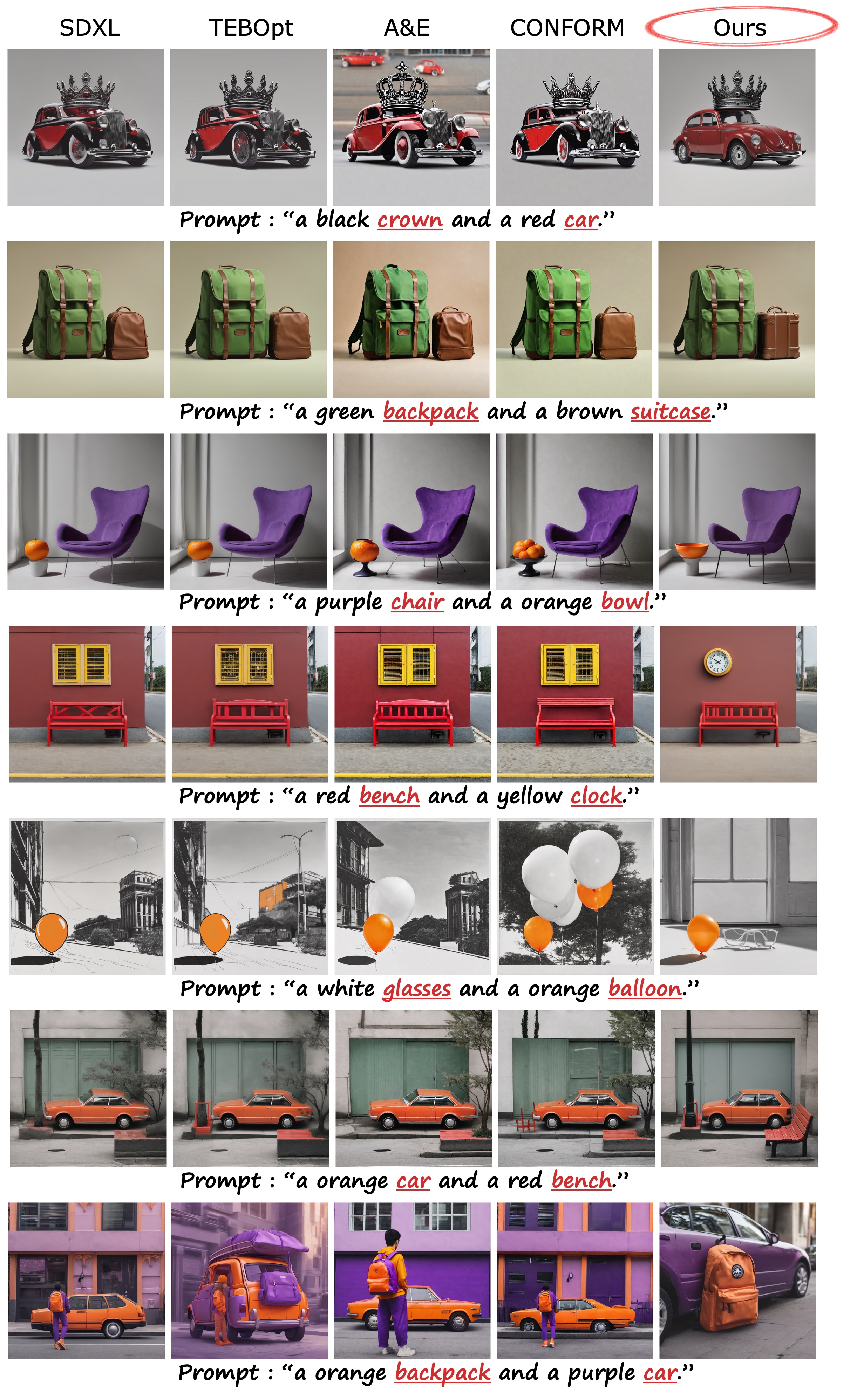}
  \captionof{figure}{Qualitative comparison on A\&E benchmark (Part 1 of 2).}
  \label{fig:qualitative_attribute_benchmark_1}
\end{figure*}

\begin{figure*}[t]
  \centering
  \includegraphics[width=0.7\textwidth]{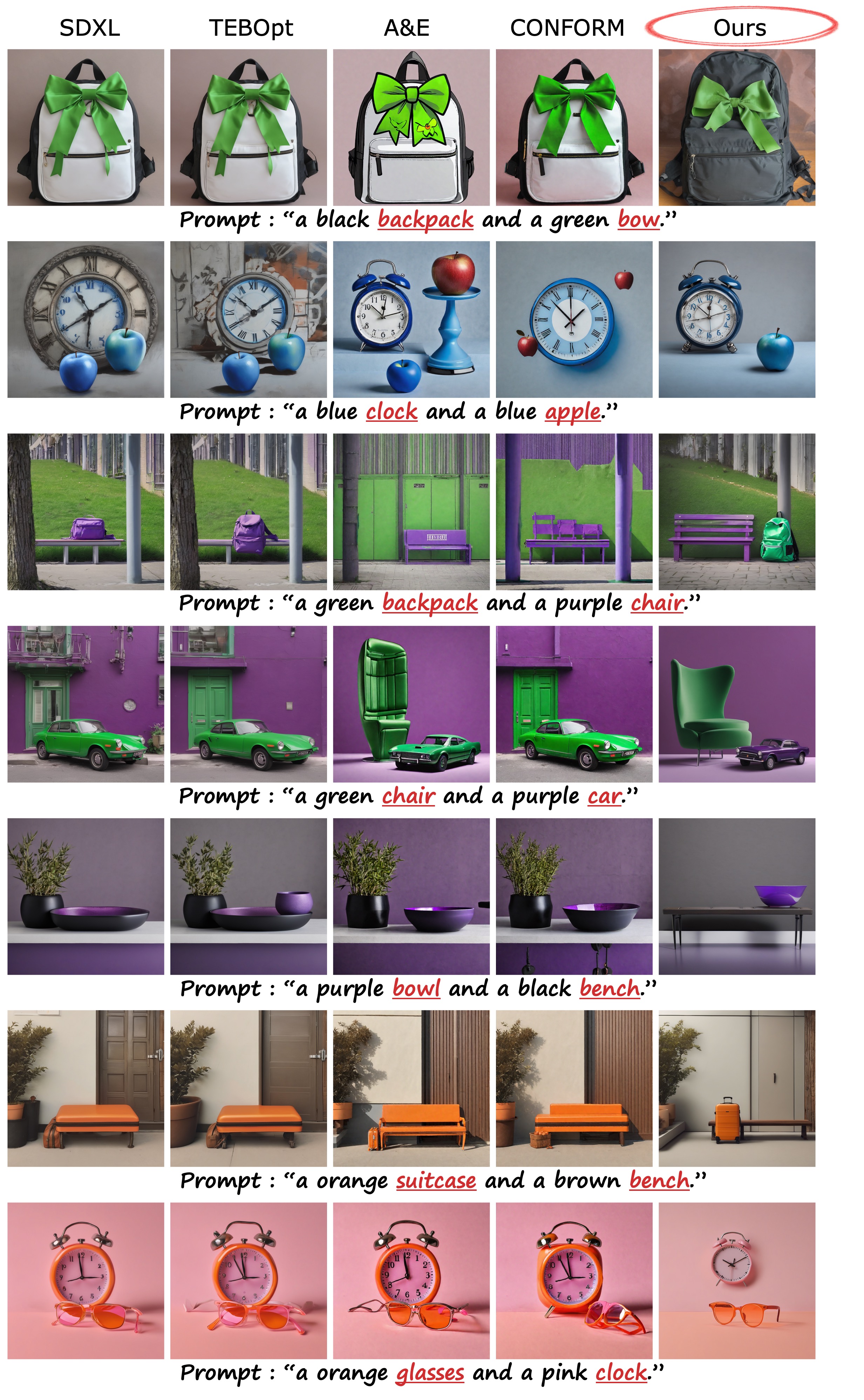}
  \captionof{figure}{Qualitative comparison on A\&E benchmark (Part 2 of 2).}
  \label{fig:qualitative_attribute_benchmark_2}
\end{figure*}

\paragraph{Qualitative results on A\&E benchmark.} Although our main benchmarks focus on failures arising from inter-object relationships in \emph{attribute‑free} prompts, we additionally demonstrate that DOS maintains and can even improve attribute binding on the widely used Attend-and-Excite (A\&E) benchmark~\cite{chefer2023attend}.
The benchmark defines three prompt templates: (i) ``a \{$animal~A$\} and a \{$animal~B$\}'', (ii) ``a \{$animal$\} and a \{$color$\} \{$object$\}'', and (iii) ``a \{$color~A$\} \{$object~A$\} and a \{$color ~B$\} \{$object~B$\}'', of which template (iii) is the most challenging because the model must correctly bind two independent color–object pairs in a single image. Figures~\ref{fig:qualitative_attribute_benchmark_1} and \ref{fig:qualitative_attribute_benchmark_2} compare DOS with four baselines on template (iii): SDXL, TEBOpt, A\&E, and CONFORM. Despite the absence of an explicit attribute–object binding guidance, DOS consistently keeps each color attached to the correct object and delivers clearer separation between the two objects, whereas competing methods often fail to bind attributes correctly or neglect the objects. Although the separation vectors in DOS were devised to mitigate inter‑object interference, they also implicitly preserve the binding between attributes and objects. These results suggest that DOS can be applied effectively to general prompts containing multiple attribute-object tuples, extending its utility beyond specific scenarios.

\begin{figure*}[t]
  \centering
  \includegraphics[width=\textwidth]{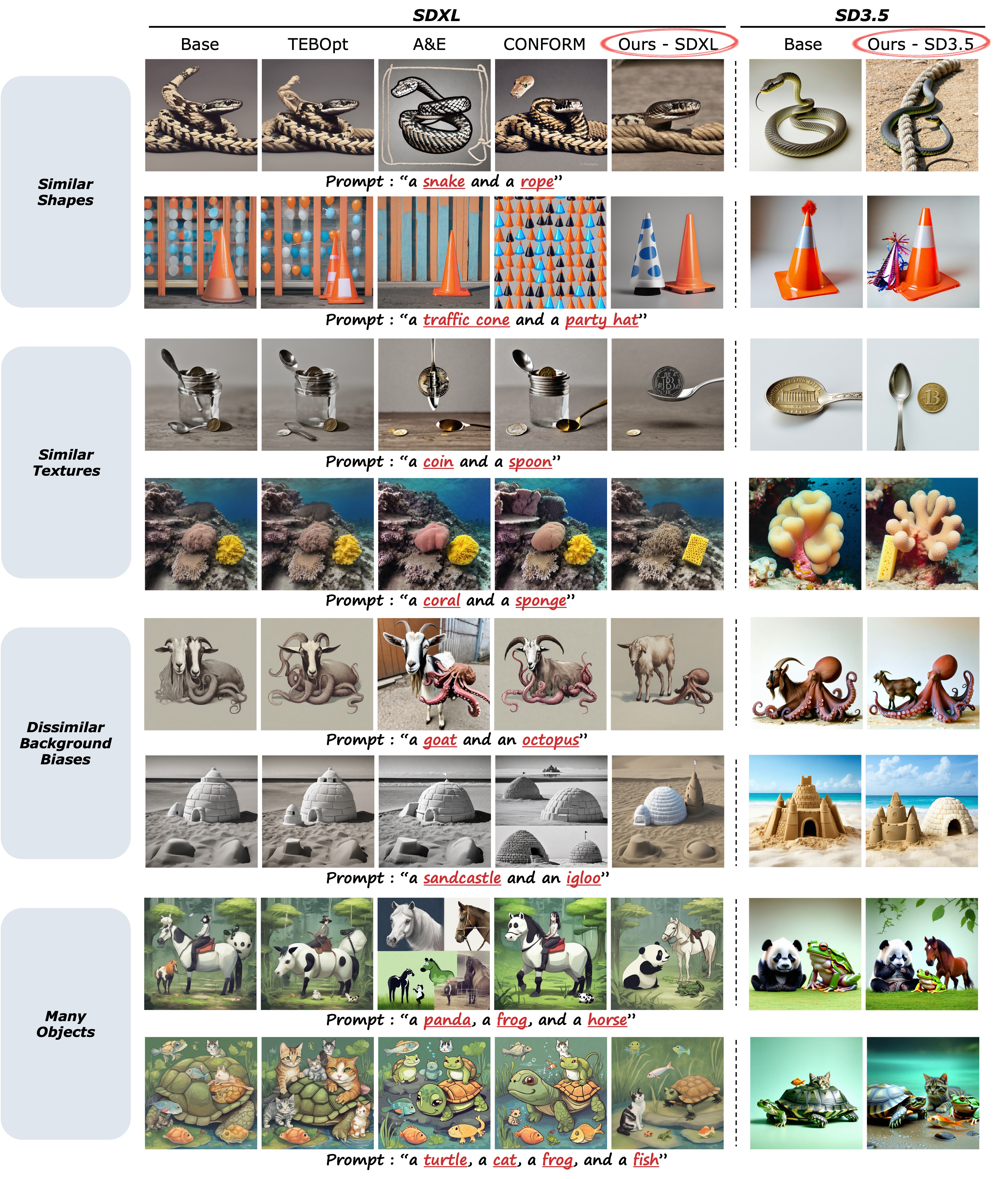}
  \captionof{figure}{Additional qualitative results on our four benchmarks (Part 1 of 3).}
  \label{fig:qualitative_extension_1}
\end{figure*}

\begin{figure*}[t]
  \centering
  \includegraphics[width=\textwidth]{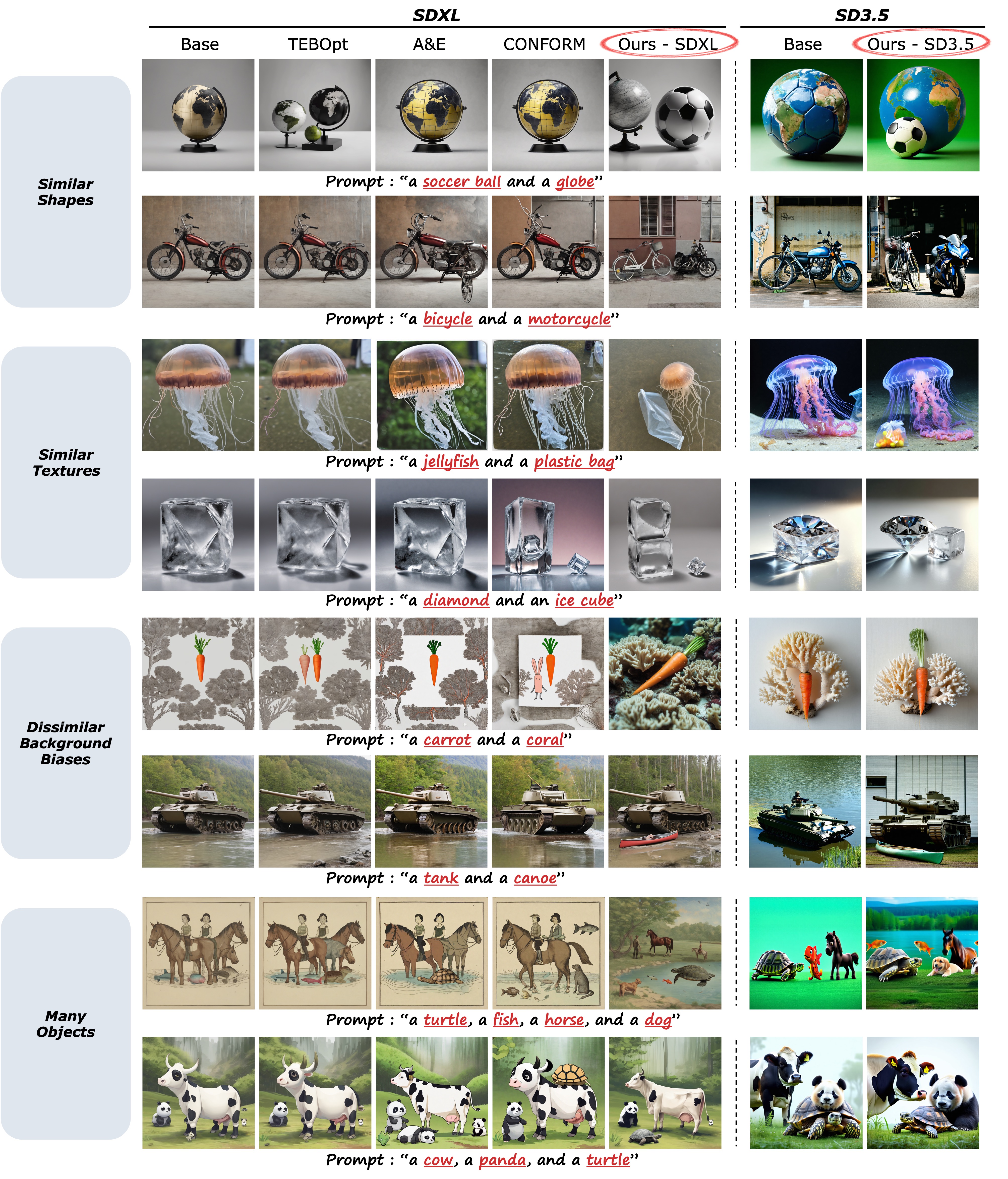}
  \captionof{figure}{Additional qualitative results on our four benchmarks (Part 2 of 3).}
  \label{fig:qualitative_extension_2}
\end{figure*}

\begin{figure*}[t]
  \centering
  \includegraphics[width=\textwidth]{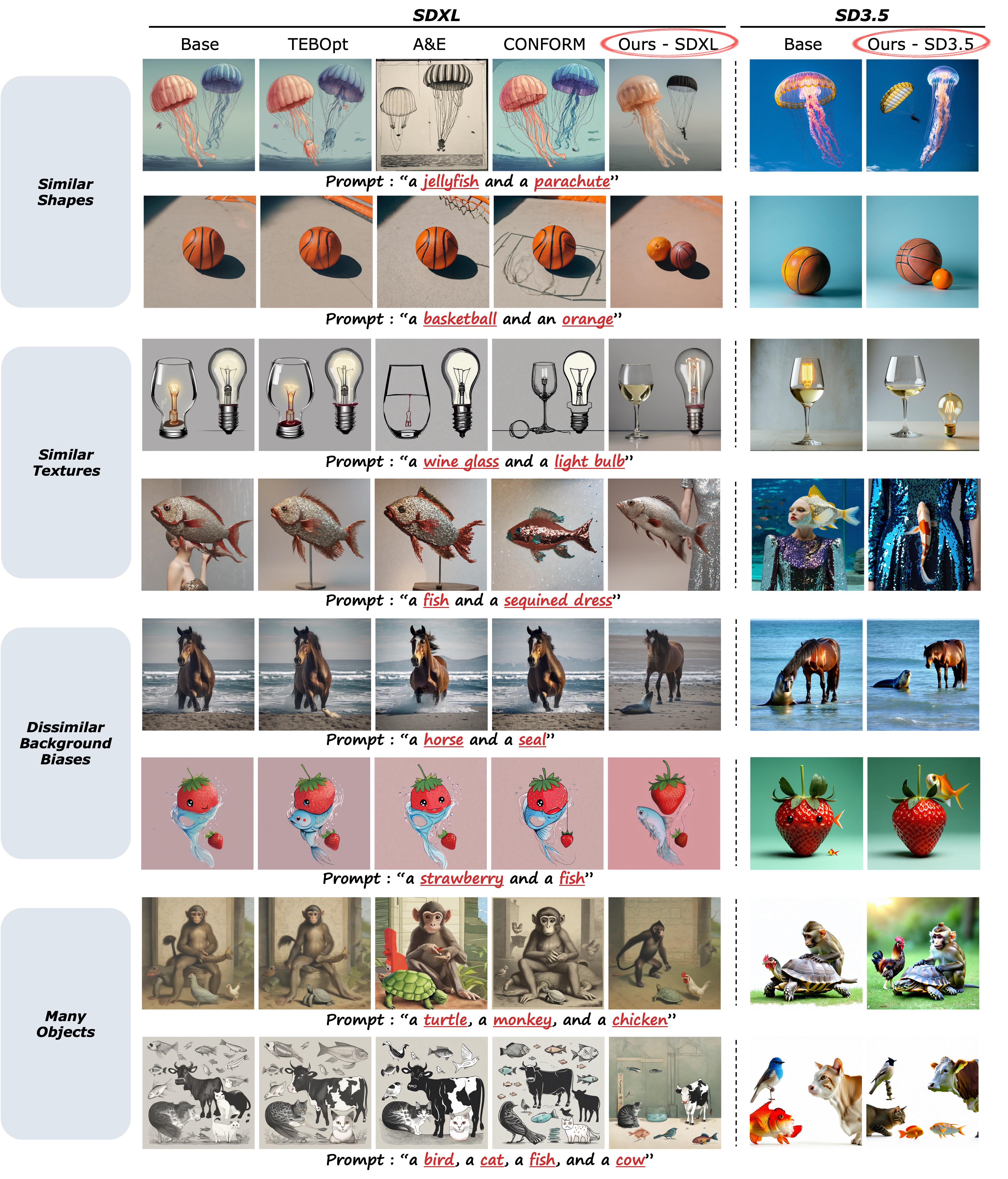}
  \captionof{figure}{Additional qualitative results on our four benchmarks (Part 3 of 3).}
  \label{fig:qualitative_extension_3}
\end{figure*}

\begin{figure*}[t]
  \centering
  \includegraphics[width=\textwidth]{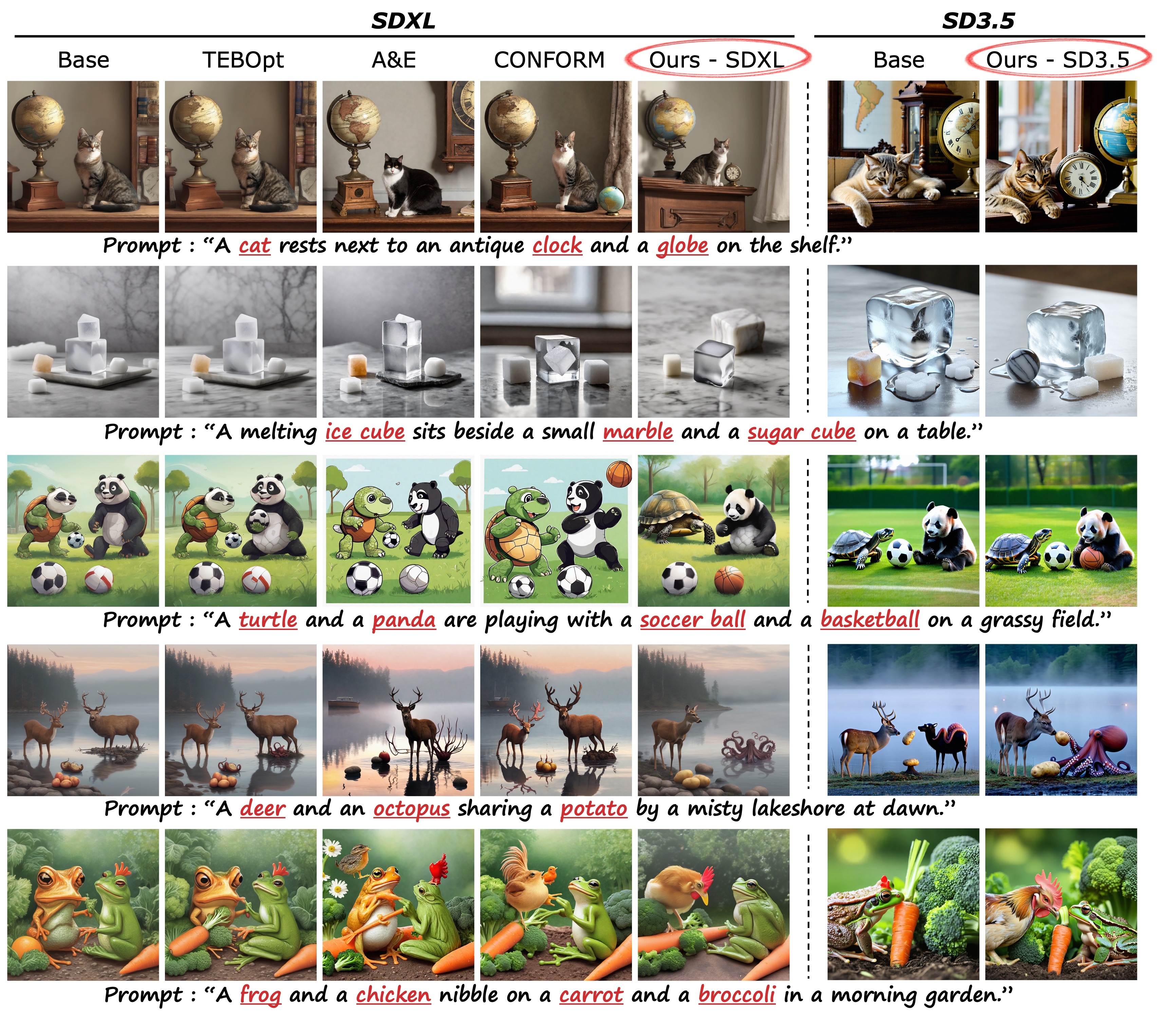}
  \captionof{figure}{Additional qualitative results on complex prompts.}
  \label{fig:general_prompt_extension}
\end{figure*}

\paragraph{Additional qualitative results.} Figures~\ref{fig:qualitative_extension_1}–\ref{fig:qualitative_extension_3} present additional qualitative results on the four benchmarks, and Figure~\ref{fig:general_prompt_extension} presents additional qualitative results on complex prompts.

\clearpage
\clearpage
\clearpage
\clearpage
\clearpage
\clearpage


\section{Directional Encoding Across CLIP Text Embedding Types}
\label{sec:appendix_word_analogy}

One of key motivations for our method is the observation that the difference between two CLIP text embeddings encodes directional information, such as ``$king - man + woman \approx queen$''~\cite{hu2024token}.
However, prior work has only validated this directional encoding when semantic token and EOT embeddings are used \emph{together}.
Since DOS calculates separation vectors \emph{independently} for different embedding types—the semantic token for each object noun and the EOT/pooled embedding—here we test whether this directional property holds for each type of embedding individually.

\begin{figure}[h]
  \centering
  \includegraphics[width=0.83\columnwidth]{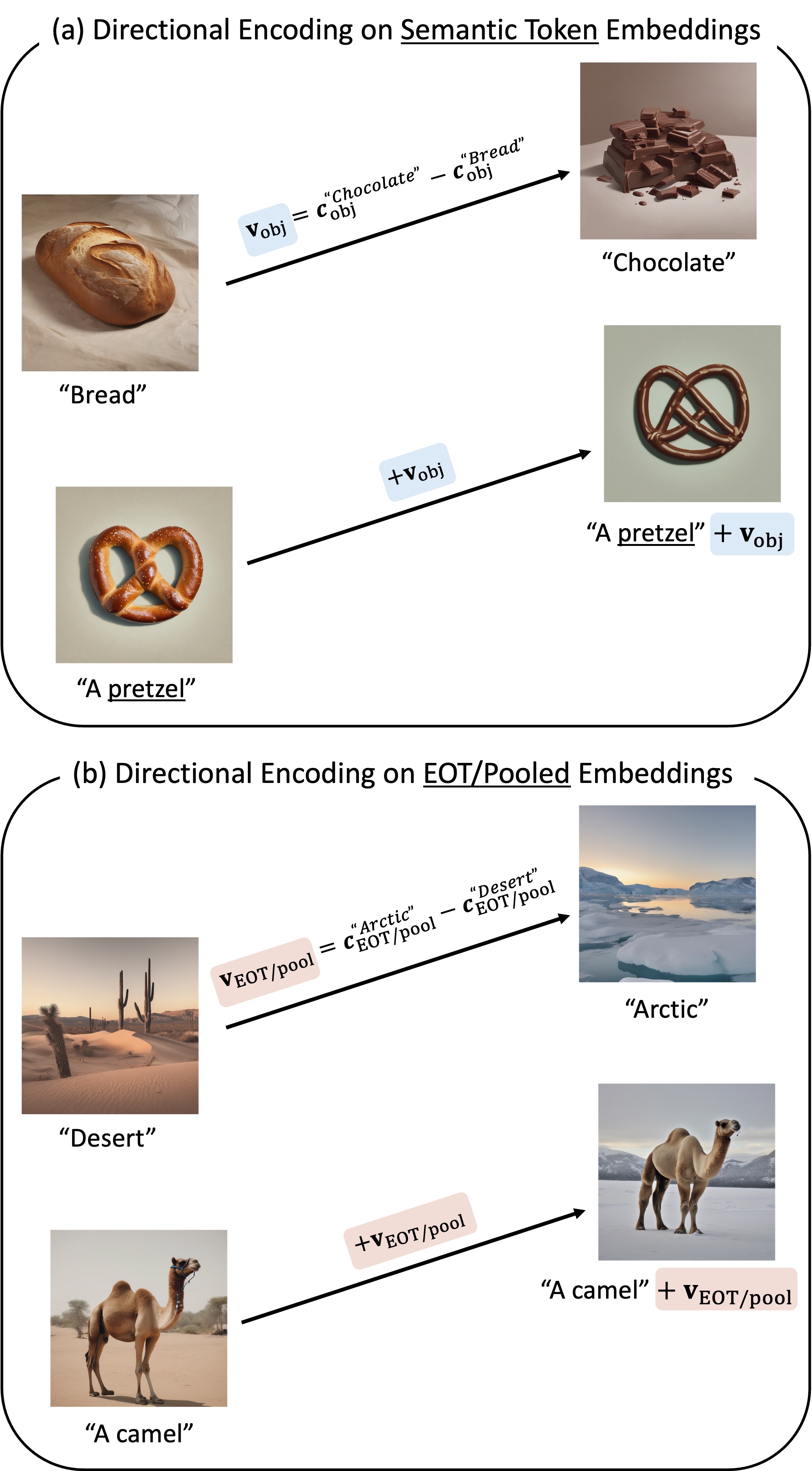}
  \captionof{figure}{Directional information encoded in the difference between two CLIP text embeddings.
  (a) \emph{Semantic token embeddings.}  
  The difference between “Chocolate’’ and “Bread’’ token embeddings, 
  $\mathbf{v}_{\text{obj}}$, is added to the “pretzel’’ token embedding, turning a bread‑like pretzel into a chocolate‑coated pretzel.  
  (b) \emph{EOT/pooled embeddings.}  
  The “Arctic–Desert’’ direction obtained from EOT/pooled embeddings, $\mathbf{v}_{\text{EOT/pool}}$, is added to the EOT/pooled embeddings of the prompt “A camel”, converting the desert background into an arctic scene while maintaining the subject.}
  \label{fig:additivity_full}
\end{figure}

Let $P$ denote a prompt containing only a single object noun~(e.g., “Bread”), and denote the semantic token embedding of this object noun by $\boldsymbol{c}^{P}_{\mathrm{obj}}\in\mathbb{R}^{d}$.  
Given two prompts $P_{1}$ and $P_{2}$, each containing a different object, we define a directional vector $\mathbf{v}_{\mathrm{obj}} \;=\; \boldsymbol{c}^{P_{2}}_{\mathrm{obj}} \;-\; \boldsymbol{c}^{P_{1}}_{\mathrm{obj}}$. Adding $\mathbf{v}_{\mathrm{obj}}$ to the semantic token embedding of a third prompt $P_{3}$ moves the generated image along the intended direction: the attribute associated with $P_1$'s object fades, whereas that of $P_2$'s object emerges.  
Figure~\ref{fig:additivity_full}(a) illustrates this behavior—adding the “chocolate–bread’’ vector to the “pretzel’’ token embedding  replaces bread‑like appearance with chocolate‑like appearance.

An analogous experiment on the EOT/pooled embeddings is performed by constructing
$\mathbf{v}_{\mathrm{EOT/pool}}=\boldsymbol{c}^{\text{``Arctic''}}_{\mathrm{EOT/pool}}-\boldsymbol{c}^{\text{``Desert''}}_{\mathrm{EOT/pool}}$
and adding it to $\boldsymbol{c}^{\text{``A camel''}}_{\mathrm{EOT/pool}}$.  
As shown in Figure~\ref{fig:additivity_full}(b), the background transitions from a desert to an arctic landscape while preserving the foreground camel.

Taken together, these results demonstrate that the directional information captured by the difference between two CLIP text embeddings can be observed independently in both the semantic token embeddings and the EOT/pooled embeddings.

\clearpage


\section{Training Data Biases Underlying Generation Failures}
\label{sec:appendix_object_co_occurrence}

The representation of the trained models are known to be significantly influenced by the distribution of their training dataset~\cite{nguyen2022quality, qin2023class, xu2024demystifying, khalafi2024constrained}. Frequent failures in the two problematic scenarios, Dissimilar Background Biases and Many Objects, can be partially explained by this effect. To investigate this, we analyze all captions in LAION 400M~\cite{schuhmann2021laion}, a dataset whose filtered subset is believed to be used in training T2I models. Specifically, we compare (i) the co-occurrence rate of object pairs with similar vs. dissimilar background biases, and (ii) the proportion of captions containing varying numbers of noun chunks. 

\begin{table}[ht]
  \centering
  \scriptsize
  \resizebox{1.0\columnwidth}{!}{%
    \begin{tabular}{@{}llcc@{}}
      \toprule
      Condition    & Object Pair               & Intersection / Union & Co-occurrence (\%) \\
      \midrule
      \multirow{11}{*}{\makecell[lt]{Similar \\ Background Biases}}
        & camel, ostrich            & 20 / 40655   & 0.049  \\
        & giraffe, elephant         & 1228 / 95385 & 1.287  \\
        & shark, whale              & 1660 / 64800 & 2.562  \\
        & koala, kangaroo           & 152 / 15641  & 0.972  \\
        & hippo, crocodile          & 76 / 26204   & 0.290  \\
        & goat, sheep               & 1026 / 85127 & 1.205  \\
        & crab, octopus             & 348 / 38095  & 0.914  \\
        & lion, tiger               & 1185 / 140775& 0.842  \\
        & penguin, polar bear       & 307 / 44326  & 0.693  \\
        & eagle, hawk               & 671 / 60597  & 1.107  \\
        \cmidrule(lr){2-4}
        & \textit{Average}          &              & 0.992  \\
      \midrule
      \multirow{11}{*}{\makecell[lt]{Dissimilar \\ Background Biases}}
        & camel, penguin            & 13 / 65264   & 0.020  \\
        & shark, elephant           & 59 / 110663  & 0.053  \\
        & horse, whale              & 71 / 190411  & 0.037  \\
        & koala, walrus             & 0 / 9243     & 0.000  \\
        & eagle, crocodile          & 8 / 69615    & 0.011  \\
        & goat, octopus             & 3 / 43116    & 0.007  \\
        & giraffe, dolphin          & 11 / 37966   & 0.029  \\
        & llama, hawk               & 0 / 17157    & 0.000  \\
        & gorilla, polar bear       & 3 / 25229    & 0.012  \\
        & kangaroo, jellyfish       & 9 / 13509    & 0.067  \\
        \cmidrule(lr){2-4}
        & \textit{Average}          &              & 0.024  \\
      \bottomrule
    \end{tabular}%
  }
  \caption{Co-occurrence rates of object pairs with similar vs. dissimilar background biases in LAION 400M.
The co-occurrence rate is computed using the Jaccard index between two objects, calculated as the number of captions containing both objects divided by the number containing at least one. Object pairs with dissimilar background biases co-occur significantly less often than those with similar biases.
}
  \label{tab:appendix_dicussion_full_table}
\end{table}

In Table~\ref{tab:appendix_dicussion_full_table}, we report the co-occurrence rates for two sets of object pairs: one with similar background biases, and the other with dissimilar biases. The co-occurrence rate is defined using the Jaccard index, calculated as the number of captions containing both objects divided by the number of captions containing at least one of them. The results show that object pairs with dissimilar background biases co-occur significantly less frequently than those with similar biases. In Table~\ref{tab:noun-chunk_distribution}, we examine how the number of noun chunks varies across prompts. Using spaCy~\cite{honnibal2020spacy} to count noun chunks per caption, we find that over 63\% of captions contain fewer than three noun chunks. Together, these findings provide evidence that the failures of T2I models in these two scenarios are, at least in part, attributable to biases in the training data distribution.

\begin{table}[h]
  \centering
  \setlength{\tabcolsep}{3.5pt}
  \scriptsize
   \resizebox{\columnwidth}{!}{%
    \begin{tabular}{@{}lcccccccc@{}}
      \toprule
      Num. of Noun Chunks & 0 & 1 & 2 & 3 & 4 & 5& $N>5$ & Total  \\
      \midrule
      Percentage~(\%)
     & 1.18 & 34.51 & 28.58 & 17.29 & 8.48 & 4.04& 5.92 & 100\\
      \bottomrule
    \end{tabular}%
  }
  \caption{Distribution of noun chunk counts in LAION 400M captions. The proportion of captions with varying numbers of noun chunks, extracted using spaCy~\cite{honnibal2020spacy}, shows that over 63\% of captions contain fewer than three noun chunks, suggesting limited multi-object descriptions in the training data.}
  \label{tab:noun-chunk_distribution}
\end{table}

\clearpage


\end{document}